\documentclass[11pt]{article}

\usepackage[top=1in, bottom=1in, left=0.75in, right=0.75in]{geometry}

\usepackage{amssymb}
\usepackage{bm}
\usepackage{amsmath}
\usepackage[ruled]{algorithm2e}
\usepackage[usenames,dvipsnames]{color}
\usepackage{array}
\usepackage{subcaption}
\usepackage{graphicx}
\usepackage{lineno}
\usepackage{tikz}
\usetikzlibrary{shapes, arrows.meta, positioning, calc}
\graphicspath{{Figures/}}
\usepackage{booktabs}
\usepackage{float}
\usepackage{caption}
\usepackage{placeins}
\usepackage{longtable}
\usepackage{tabularx}
\usepackage{enumitem}
\usepackage[numbers,sort&compress]{natbib}

\usepackage{hyperref}
\hypersetup{
    colorlinks,
    linkcolor={magenta},
    citecolor={blue},
    urlcolor={blue!80!black},
    breaklinks=true,
    plainpages=true
}

\newcommand{\cref}[3]{\hyperref[#2]{#1~\ref*{#2}{#3}}}
\newcommand{\crefs}[3]{\hyperref[#2]{#1~\ref*{#2}-\ref*{#3}}}
\newcommand{\colref}[2]{\hyperref[#2]{#1~\ref*{#2}}}
\newcommand{\eqnref}[1]{\colref{Eq.}{#1}}
\newcommand{\figref}[1]{\colref{Figure}{#1}}

\newcommand{\secref}[1]{\colref{Section}{#1}}

\newcommand{\tabref}[1]{\colref{Table}{#1}}
\newcommand{\algref}[1]{\colref{Algorithm}{#1}}

\newcommand{\doi}[1]{\textsc{doi}: \href{http://dx.doi.org/#1}{\nolinkurl{#1}}}

\title{FloraForge: LLM-Assisted Procedural Generation of Editable and Analysis-Ready 3D Plant Geometric Models For Agricultural Applications}

\author{
  Mozhgan Hadadi\textsuperscript{1},
  Talukder Z. Jubery\textsuperscript{1},
  Patrick S. Schnable\textsuperscript{1},
  Arti Singh\textsuperscript{1},
  Bedrich Benes\textsuperscript{2},\\
  Adarsh Krishnamurthy\textsuperscript{1,*},
  Baskar Ganapathysubramanian\textsuperscript{1,*}
}

\date{
  \textsuperscript{1}Iowa State University, Ames, IA 50011, USA\\
  \textsuperscript{2}Purdue University, West Lafayette, IN 47906, USA\\[1ex]
  \textsuperscript{*}\,Corresponding authors
}

\begin{document}
\maketitle

\begin{abstract}
Accurate three-dimensional plant models are crucial for computational phenotyping and physics-based simulation; however, current approaches face significant limitations. Learning-based reconstruction methods require extensive species-specific training data and lack editability for hypothesis-driven research. Procedural modeling offers parametric control and large model variability but demands specialized expertise in geometric modeling and an in-depth understanding of complex procedural rules, making it inaccessible to domain scientists. We present FloraForge, an LLM-assisted framework that enables domain experts to generate biologically accurate, fully parametric 3D plant models through iterative natural language Plant Refinements (PR), minimizing the need for programming expertise. Our vibe coding system leverages LLM-enabled co-design to progressively refine Python scripts that generate parameterized complex plant geometries as hierarchical B-spline surface representations with botanical constraints. Plant organs are represented as B-spline surfaces with explicit control points and parametric deformation functions. This representation can be easily tessellated into polygonal meshes with arbitrary precision, ensuring compatibility with functional structural plant analysis workflows such as light simulation, computational fluid dynamics, and finite element analysis. We demonstrate the framework on maize (5 genotypes), soybean, and mung bean (6 genotypes), fitting procedural models to empirical point cloud data through manual refinement of the Plant Descriptor (PD), human-readable YAML files that the LLM originally templated. The pipeline generates dual outputs: triangular meshes (represented as STL or OBJ files) for visualization and triangular meshes with additional parametric metadata for quantitative analysis (stored as SMESH files). This approach uniquely combines pre-trained LLM-assisted template creation, mathematically continuous representations enabling both phenotyping and rendering, and direct parametric control through PD. The framework democratizes sophisticated geometric modeling for plant science while maintaining mathematical rigor through biologically interpretable parameterizations. Moreover, the PR dialogue leading to the 3D model captures an in-depth understanding of the properties of the generated model, a step that is often missing in procedural modeling. 
\end{abstract}


\section{Introduction}
\label{sec:Intro}

Accurate three-dimensional (3D) plant geometric models are foundational to modern plant science, enabling computational phenotyping, physics-based simulation of agroecosystems, and functional-structural modeling of growth processes~\citep{paulus2019measuring,gibbs2016approaches,vos2010functional,louarn2020two}. However, generating such models from scratch (forward modeling) or finding a 3D model representation for captured data (reconstruction) remains a significant bottleneck: learning-based reconstruction methods require extensive species-specific training datasets and produce representations that resist direct editing, while procedural modeling systems demand specialized expertise in geometric programming and rule-based algorithms that few plant scientists possess. So-called inverse procedural models often represent only a class of objects and fail on detailed geometries~\cite{guo2020inverse,Stava14CFG,Stava10CGF}. This fundamental tension between accessibility and analytical utility has constrained the adoption of sophisticated 3D modeling in plant phenotyping, breeding, and agronomic research.

Current approaches face three fundamental limitations that prevent widespread adoption. First, learning-based reconstructions~\citep{cheng2025demeter,yang2025neuraleaf,liu2025neural} yield mesh or latent representations that lack parametric editability, thereby preventing researchers from modifying biologically meaningful traits, such as leaflet pitch, internode elongation, or organ curvature, for hypothesis-driven studies. Second, procedural modeling frameworks such as L-systems~\citep{prusinkiewicz2012algorithmic,prusinkiewicz1999system} and template-based systems~\citep{beardsley2017editable,raistrick2023infinite} require deep expertise in programming the grammar rules and geometric modeling, making them inaccessible to domain scientists who lack specialized training in 3D graphics or geometric modeling. Moreover, existing procedural models are complex, non-linear systems that are difficult to control. 
Third, most existing reconstruction methods are suited for artificial objects, such as CAD models (e.g., \cite{li2023secad,Li24CVPRa}), and fail on thin and long geometries, which is a typical case for vegetation. These limitations force researchers to choose between accessible but uneditable and approximate reconstructions and powerful yet inaccessible procedural tools, a choice that ultimately hinders scientific progress.

\begin{figure}[t!]
    \centering
    \includegraphics[trim=0cm 0cm 0cm 0cm, clip, width=0.8\linewidth]{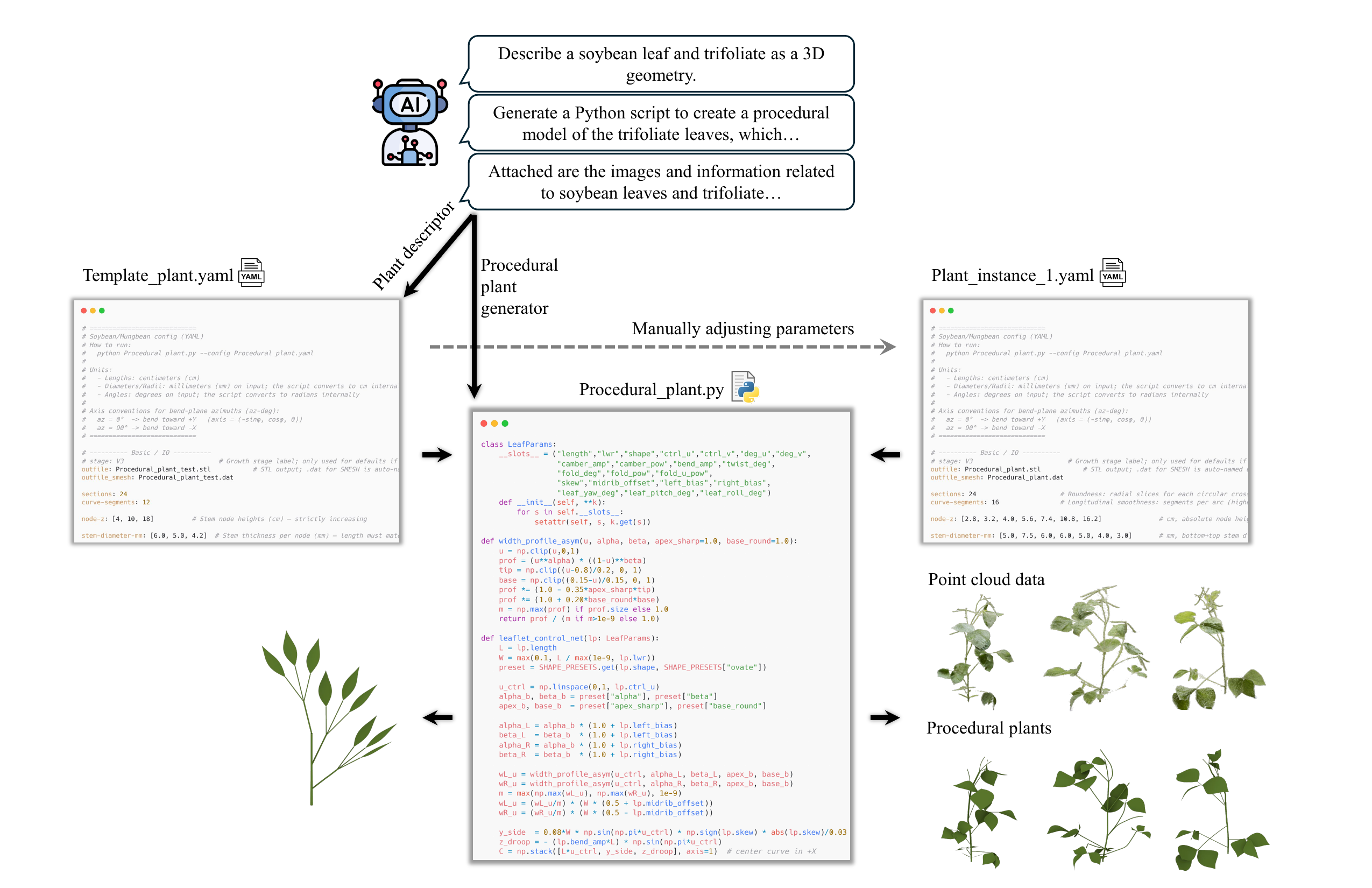}
    \caption{\textbf{FloraForge}: LLM-assisted vibe procedural modeling workflow. Users provide Plant Refinements (PR) to obtain a Procedural Plant Generator (PPG; Python) and a Plant Descriptor (PD; YAML). Executing PPG(PD) yields an Initial 3D Template (IT) with full parametric control, allowing domain experts without programming expertise to develop custom modeling tools through iterative refinement.}
    \label{fig:pipeline}
\end{figure}

Recent advances in large language models (LLMs) have demonstrated exceptional capabilities in code synthesis, multimodal reasoning, and tool utilization for 3D content creation~\citep{sun20253d,lu2025ll3m,du2024blenderllm,jignasu2024evaluating}. These models demonstrate remarkable proficiency in generating correct object attributes, parsing textual descriptions, correctly interpreting code functions, and facilitating efficient human-computer interactions through natural language dialog. Despite these advances, existing LLM-driven geometric modeling systems primarily focus on general objects, such as CAD models or architectural applications, without addressing the unique challenges of botanical morphology, including hierarchical organ structures, species-specific features, and growth constraints, phyllotactic arrangements, and the need for biologically interpretable parameterizations that directly map to phenotypic traits. Moreover, these systems typically generate assets for visualization rather than scientific analysis, usually lacking the mathematical rigor required for quantitative phenotyping and computational simulation. A critical gap lies at the intersection of accessibility and analytical rigor. Plant scientists need tools that are as accessible as learning-based reconstruction methods yet provide the parametric control and mathematical continuity of procedural modeling, without requiring programming expertise or species-specific training data. Bridging this gap requires three synergistic capabilities: (1)~automated generation of procedural templates incorporating botanical domain knowledge, (2)~mathematically continuous parametric geometric representations suitable for both quantitative analysis and realistic visualization, and (3)~human-readable high-level biologically meaningful parameter interfaces enabling direct editing by domain experts.

We present \textbf{FloraForge} (see \figref{fig:pipeline}), an LLM-assisted framework that bridges this gap through three synergistic innovations. First, we leverage LLMs to automate the creation of procedural modeling templates from botanical descriptions, eliminating the need for manual geometric programming. Through structured iterative dialog, domain experts progressively refine Python scripts that implement continuous hierarchical B-spline surface representations with botanical constraints, without requiring expertise in geometric modeling. Second, we represent plant organs as continuous B-spline surfaces that provide both the mathematical properties required for quantitative analysis (smoothness, differentiability, and exact curvature computation) and the realism needed for visualization. The Non-Uniform Rational B-spline (NURBS) representations ensure compatibility with a variety of analysis pipelines \footnote{A key requirement for performing high fidelity computational analysis is accurate 3D geometric representation. However, this is not the only requirement. One often needs associated material properties to perform these analyses. These could include surface properties (e.g., BRDF) for radiation simulations and mechanical properties (e.g., stiffness) for structural simulations.} -- radiation simulations, computational fluid dynamics, and finite element workflows -- while supporting physically realistic rendering. Third, we expose all morphological parameters through Plant Descriptor (PD) (human-readable YAML files), enabling domain experts to directly edit and version-control plant architectures without programming knowledge. Hierarchical parameter inheritance (global defaults with node-specific and organ-specific overrides) minimizes redundancy while preserving biological realism and flexibility. This combination of LLM-assisted template generation, mathematically rigorous representations, and accessible parametric control uniquely positions our framework to democratize sophisticated geometric modeling for plant science. 
FloraForge lowers the barrier for domain scientists to prototype custom modeling tools for the plant forms studied here, supporting the iterative refinement of both procedural algorithms (via natural-language dialogue with the LLM) and individual plant instances (via PD parameter editing). We illustrate on monocot and dicot crop architectures, with natural future extensions to vines, shrubs, and trees.

Our framework addresses the critical limitations of existing approaches through four key differentiators. First, compared to learning-based reconstruction methods such as Demeter~\citep{cheng2025demeter} (which require 600+ annotated scans) and NeuraLeaf~\citep{yang2025neuraleaf} (which require thousands of 2D leaf samples per species), our approach is \textit{training-free}, utilizing existing pre-trained LLMs through iterative natural language dialog, generating explicit, biologically interpretable parameters stored in version-controllable YAML files rather than opaque latent vectors. This eliminates the data acquisition bottleneck and provides the transparency essential for scientific reproducibility. Second, unlike traditional L-systems~\citep{prusinkiewicz2012algorithmic,deussen1998realistic,mvech1996visual} that represent plants as discrete branching graphs requiring expert rule design, we provide continuous B-spline surface representations with guaranteed smoothness ($C^2$ continuity) for precise morphometric measurements, curvature analysis, and numerical simulations. Third, in contrast to inverse procedural approaches~\citep{stava2014inverse,zhai2024cropcraft} that optimize parameters for pre-existing templates, we \textit{automate the template design process itself} through LLM co-design, enabling rapid adaptation to new species without requiring manual geometric specification or optimization convergence. Finally, while recent LLM-based 3D generation systems~\citep{sun20253d,lu2025ll3m} demonstrate impressive capabilities for buildings and general objects, they lack the botanical domain knowledge and phenotyping-oriented parameterizations essential for plant science applications.

Our contributions advance the state of the art in plant geometric modeling:
\begin{itemize}[itemsep=0pt,topsep=0pt]
\item \textbf{LLM-assisted procedural plant generator:} A framework that automates the creation of bota\-nically-constrai\-ned B-spline surface templates through iterative natural language PR, eliminating the need for expertise in geometric modeling or procedural modeling systems.

\item \textbf{Analysis-ready continuous representations:} Explicit B-spline surface formulations provide mathematical continuity for precise phenotypic measurements, numerical simulation compatibility, and physically realistic rendering within a unified representation.

\item \textbf{Democratized parametric control:} Human-readable PD expose morphological parameters (curvature, width profiles, orientation, and deformation) for direct editing by domain experts, with hierarchical parameter inheritance minimizing redundancy while preserving biological realism.

\item \textbf{Cross-species validation:} Demonstration of template-based modeling across monocot (maize) and dicot (soybean, mung bean) architectures, fitting procedural models to empirical LiDAR point clouds and showcasing generalization capabilities for diverse plant morphologies and developmental stages.
\end{itemize}

The practical impact of this framework extends beyond technical capabilities. By enabling domain scientists to develop custom plant modeling tools without specialized computational expertise, FloraForge accelerates the research cycle for plant phenotyping, breeding, and precision agriculture applications. Researchers can rapidly prototype species-specific templates, fit models to acquired real-world data through intuitive parameter adjustment, and generate analysis-ready outputs compatible with downstream computational workflows—all without needing to write code or master complex procedural or geometric modeling systems.

The following sections position our work within the broader landscape of plant modeling research (\secref{sec:related_work}), detail the LLM-assisted development workflow and procedural generation algorithms (\secref{sec:methods}), and demonstrate the framework's capabilities through fitted models spanning multiple crop species (\secref{sec:results}). We conclude with a discussion of limitations, future directions, and the broader implications of LLM-assisted scientific code development for computational plant science (\secref{sec:conclusion}).

\section{Related Work}
\label{sec:related_work}

\paragraph{Procedural Plant Modeling}

Procedural modeling has long been a cornerstone of plant representation in computer graphics and computational biology. Lindenmayer's systems (L-systems), introduced by Lindenmayer~\citep{lindenmayer1968mathematical}, provide a formal grammar for modeling plant development through parallel string rewriting. These systems have evolved to incorporate stochastic variations~\citep{prusinkiewicz1994synthetic}, context-sensitive rules~\citep{prusinkiewicz1999system}, and parametric control~\citep{prusinkiewicz1986graphical}. Recent implementations, such as Infinigen~\citep{raistrick2023infinite}, demonstrate the power of procedural generation in creating diverse, realistic object geometries with adjustable parameters and rule-based systems. Despite their expressiveness, L-systems face several limitations for accessible plant modeling. First, they require large datasets for training, as well as expert knowledge to design appropriate production rules and parameter sets for new species~\citep{guo2020inverse,Lee24ToG}. Second, L-systems primarily generate branching structures represented as skeletal graphs, requiring additional algorithms to construct continuous surface meshes~\citep{hadadi2025procedural}. 
Alternative procedural approaches have explored template-based modeling~\citep{beardsley2017editable,makowski2019synthetic}, where parametric surface primitives are assembled to construct plant organs. These methods benefit from explicit geometric control but typically require manual specification of component relationships and parameter values. Recent ML-based approaches use dictionaries of templates to reconstruct 3D model of a forest~\cite{Zhou25TRGS}. Our work builds upon this foundation by automating template creation through LLM-assisted design while maintaining the editability advantages of parametric representations.

\paragraph{Inverse Procedural Modeling} 

Inverse procedural modeling (IPM)~\cite{Aliaga16SiggCourse} addresses the challenge of inferring procedural parameters from existing geometry, enabling compression and variation generation~\citep{stava2014inverse,Stava10CGF,mcquillan2018algorithms,lotfi2024optimal}. For plant modeling, IPM approaches have explored fitting L-system parameters to tree point clouds~\citep{guo2020inverse}, reconstructing branching structures from images~\citep{li2021learning}, and optimizing parametric models to match observations~\citep{zhai2024cropcraft}.
CropCraft~\citep{zhai2024cropcraft} represents recent progress in inverse procedural modeling for agricultural crops, using Bayesian optimization to fit parametric plant models to observations from neural radiance fields. While effective for reconstruction, these optimization-based approaches require predefined procedural templates and struggle to generalize to new species with different morphological characteristics.

\paragraph{Learning-Based Plant Reconstruction}

Recent advances in neural representations have enabled powerful learning-based approaches for 3D plant reconstruction. Methods like Demeter~\citep{cheng2025demeter} learn parametric morphological models from real-world data, encoding topology, articulation, shape, and deformation into compact latent spaces. TreeStructor~\cite{Lee25ECCV} builds 3D simulation-ready models of trees from single Google street-view photographs via diffusion. NeuraLeaf~\citep{yang2025neuraleaf} introduces neural parametric models that disentangle 2D base shapes from 3D deformations, enabling the reconstruction of individual leaves with natural deformations. Neural hierarchical decomposition~\citep{liu2025neural} decomposes plants into box hierarchies at varying levels of detail, accommodating both houseplants and outdoor trees. While these learning-based methods achieve impressive reconstruction quality, they present several limitations for accessible plant modeling workflows. First, they require substantial species-specific training data. Demeter collected over 600 annotated soybean scans, TreeStructor uses thousands of synthetic tree models, and NeuraLeaf requires 2D leaf datasets with thousands of samples per species~\citep{cheng2025demeter,yang2025neuraleaf}. Second, the learned representations, while compact, lack the interpretability and direct editability that domain scientists require for hypothesis-driven research. Even if some of the parameters are available, they are challenging for domain experts to comprehend. Third, these methods focus primarily on reconstruction from observations rather than generative modeling for custom species design.

Point clouds are becoming the primary method for capturing 3D geometries because of the proliferation of acquisition sensors. The reconstruction methods~\citep{hadadi2025procedural,ando2021robust} address the challenge of fitting parametric models to 3D scan data. Recent work combines NURBS surface fitting with optimization strategies~\citep{hadadi2025procedural}, achieving high-fidelity reconstruction of maize plants through a two-stage optimization process. However, these approaches assume the availability of dense 3D point clouds and focus on inverse modeling rather than forward template creation.

\paragraph{Parametric Geometric Representations}

Non-Uniform Rational B-Splines (NURBS) provide a mathematically rigorous framework for representing curves and surfaces with high precision and flexibility~\citep{farin2014curves}. NURBS are a de facto industrial standard in CAD modeling, and they offer several advantages for plant modeling: (1) exact representation of both analytic shapes and freeform surfaces, (2) local control through weighted control points, (3) mathematical continuity properties essential for quantitative analysis, and (4) compact representation compared to polygon meshes~\citep{piegl2012nurbs}. Rational B-splines are a subset of NURBS, use piecewise polynomial functions to define smooth curves and surfaces~\citep{catmull1974class}. The local control property, where changes to individual control points and their knots affect only nearby regions, makes B-splines particularly suitable for modeling organic shapes with localized variations. Recent work has explored the use of B-splines~\citep{yang2023curve} and parametric leaf surfaces~\citep{bradley2013image}, demonstrating their effectiveness in capturing botanical morphology. Traditional approaches to NURBS-based plant modeling require manual specification of control points, knot vectors, and surface parameterizations~\citep{beardsley2017editable}. Our work addresses this limitation by leveraging LLMs to automate the design process, generating appropriate NURBS parameterizations from high-level botanical descriptions while maintaining the mathematical properties essential for scientific applications.

\paragraph{LLM-Assisted 3D Modeling}

LLMs have recently been explored for 3D content generation through code synthesis. 3D-GPT~\citep{sun20253d} employs multi-agent systems to break down procedural modeling tasks, generate Python code, and interface with Blender for scene generation. LL3M~\citep{lu2025ll3m} presents a multi-agent system that writes interpretable Python code for creating 3D assets, emphasizing modularity and integration with artist workflows. BlenderLLM~\citep{du2024blenderllm} fine-tunes language models specifically for CAD script generation based on user instructions. These systems demonstrate the potential of LLMs for democratizing 3D content creation. However, they focus primarily on general geometric modeling of manufactured objects (such as chairs, airplanes, cups, and lamps) or architectural applications, lacking the botanical knowledge and domain-specific constraints necessary for plant modeling. Furthermore, existing LLM-based 3D generation systems typically operate at the level of individual objects or scenes, without addressing the hierarchical organ structures, growth patterns, and phenotypic constraints that characterize plant morphology. Recent work on LLM-assisted CAD generation~\citep{wu2023cad,schupbach2025text} explores frameworks for automated geometry generation through function calling and agent workflows. These approaches enable LLMs to interpret design requirements and generate executable code for CAD software. However, they have not been applied to the unique challenges of botanical modeling, where biological priors, morphological constraints, and phenotypic parameters must be incorporated into the generation process.

Our work takes a complementary approach: rather than optimizing parameters for pre-existing templates, we automate the template creation process itself. By leveraging LLMs to design appropriate parametric representations from botanical descriptions, we enable rapid prototyping of modeling tools for diverse plant species without requiring manual template engineering or extensive optimization. FloraForge occupies a unique position at the intersection of procedural modeling, parametric representations, and LLM-assisted design. Unlike learning-based methods that require extensive training data, we exploit the reasoning capabilities of pre-trained LLMs. Unlike traditional procedural systems that demand expert knowledge, we provide an accessible interface through natural language descriptions and human-readable parameter files. Unlike existing LLM-based 3D generation systems, we incorporate botanical knowledge and domain-specific constraints essential for plant phenotyping applications. Our system also does not provide descriptions of low-level geometric parameters; instead, it offers more meaningful biological parameters. The combination of LLM-assisted template design, explicit B-spline representations, and human-editable parameters addresses the critical gap in accessible, editable plant modeling tools for domain scientists.

\section{Methods}
\label{sec:methods}

We have developed an editable, analysis-ready procedural modeling pipeline for generating biologically accurate 3D plant architectures. The system integrates LLM-assisted iterative co-design, parametric B-spline surface representations, and hierarchical procedural architectural frameworks to produce geometrically precise, fully parametric plant models. We demonstrate it on both dicotyledonous species (soybean \textit{Glycine max}, mung bean \textit{Vigna radiata}) and monocotyledonous species (maize \textit{Zea mays}). All procedural parameters are stored in Plant Descriptors (PDs), ensuring complete reproducibility, version control, and editability of generated plant architectures.

\begin{figure}[hbt]
    \centering
    \begin{subfigure}[t]{0.49\linewidth}
        \centering
        \includegraphics[trim=0cm 0cm 0cm 1.5cm, clip, width=\linewidth]{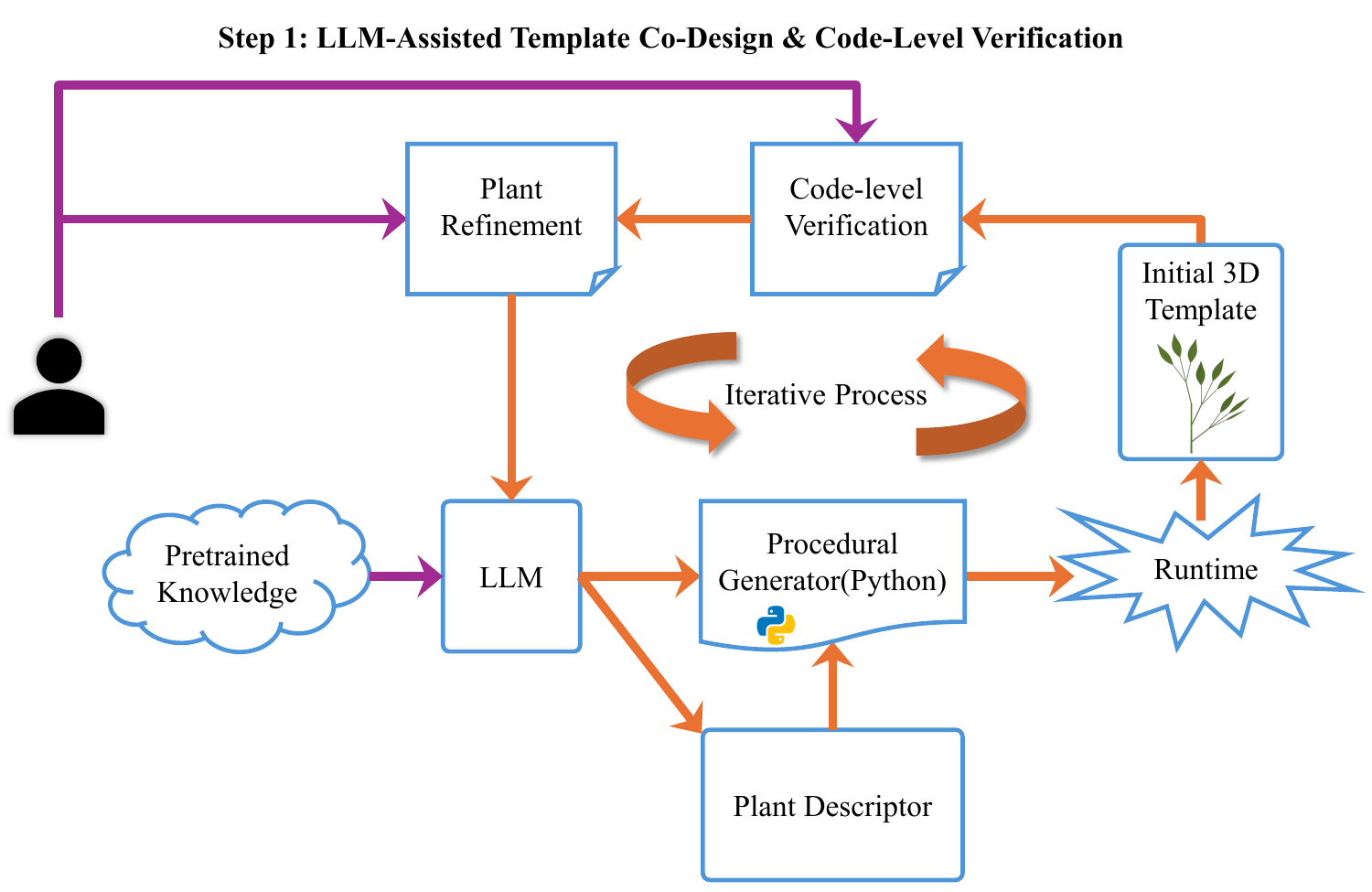}
        \caption{Step 1: LLM-Assisted Template Co-Design \& Code-Level Verification
}
        \label{fig:twostep_s1}
    \end{subfigure}\hfill
    \begin{subfigure}[t]{0.49\linewidth}
        \centering
        \includegraphics[trim=0cm 0cm 0cm 1.5cm, clip, width=\linewidth]{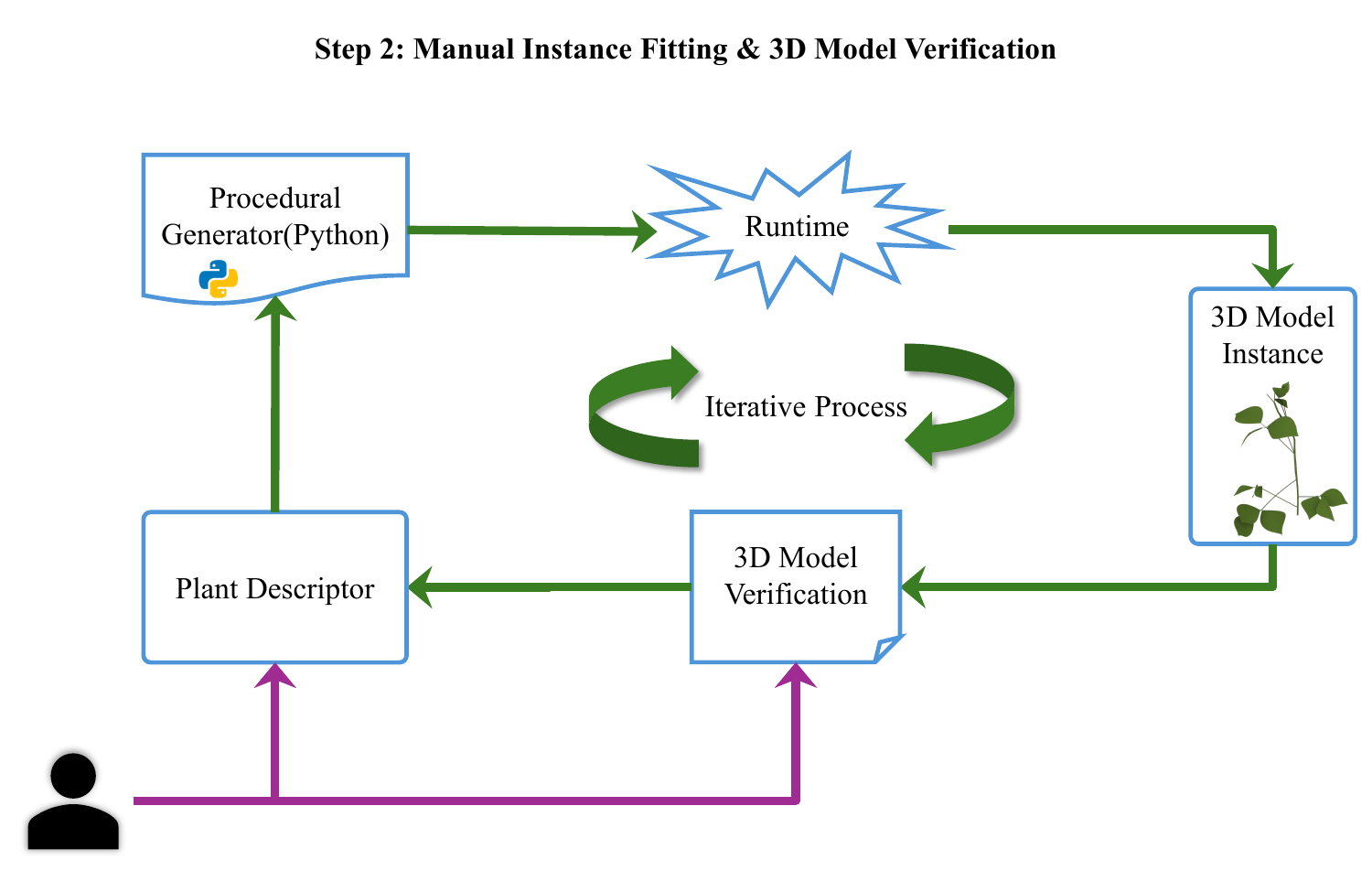}
        \caption{Step 2: Manual Instance Fitting \& 3D Model Verification
}
        \label{fig:twostep_s2}
    \end{subfigure}
    \caption{\textbf{Two-stage FloraForge workflow.} (a) LLM produces the \emph{Procedural Generator} and an \emph{LLM-seeded descriptor}; executing these yields the \emph{Initial 3D Template} and passes \emph{Code-level 
Verification}. (b) The \emph{Procedural 
Generator} is reused while the \emph{Plant Descriptor} is manually refined; executing these yields the \emph{3D Model 
Instance}, which is validated by \emph{3D Model 
Verification} (visual inspection + Chamfer distance).}

    \label{fig:twostep}
\end{figure}

The workflow comprises two stages (see \figref{fig:twostep}). In \emph{Step 1} (LLM-assisted template co-design and Code-level Verification; CLV), the user interacts with the LLM only through Plant Refinements (PR). From these prompts, the LLM synthesizes a Procedural Plant Generator (PPG; Python) with an LLM-seeded Plant Descriptor (PD). Executing the PPG with this PD yields an Initial 3D Template (IT). CLV consists of running the script and visually inspecting the output to ensure that the code and the produced geometry align with the intended modeling purpose (e.g., correct organ types and hierarchy, plausible scales/units, expected attachment logic); CLV is qualitative and does not involve quantitative fitting at this stage. In \emph{Step 2} (manual instance fitting and 3D Model Verification, MV), the PPG is held fixed while the PD is iteratively edited by the user. After each edit, the PPG is executed to render a candidate model that is visually compared against the target point cloud; edits continue until visual agreement is achieved. The final output is a Model Instance (MI), for which MV additionally quantifies agreement using the symmetric Chamfer distance (CD).

\subsection{Hierarchical Plant Architecture}
\label{sec:architecture}

Plant topology is encoded as a node-based hierarchy reflecting species-specific botanical organization. The plant geometry is generated from the topological graph by providing geometric descriptors at each node. The procedural generation framework accommodates two fundamental architectural patterns:

\paragraph*{Dicotyledonous Architecture (Soybean, Mung Bean)}

Dicot models implement a four-level hierarchy:
\begin{enumerate}
    \item \textbf{Main stalk}: Central stem with nodes positioned at absolute heights \(\{z_1, z_2, \ldots, z_N\}\) and internodes connecting consecutive nodes. Stem curvature is controlled via per-node curvature parameters.
    \item \textbf{Petioles}: Branch structures extending from nodes at species-specific azimuth, pitch, and roll angles. Each node can support one or more petioles (typically 1-2 per node in our implementations).
    \item \textbf{Petiolules}: Secondary branches extending from petiole tips in a trifoliate arrangement (terminal, left lateral, and right lateral), each with independent orientation and curvature parameters.
    \item \textbf{Leaflets}: B-spline surface leaflets attached to petiolule tips, with per-leaflet control over morphology (length, width, shape, camber, twist, fold) and orientation (yaw, pitch, roll).
\end{enumerate}
Each component exhibits distinct geometric properties: stems, petioles, and petiolules are modeled as a generalized cylinder with the main axis being the stem and the sweeping profile formed by a circular cross-section B-spline curve with controllable diameter variation and 3D curvature; leaflets are modeled as tensor product bi-variate B-spline surfaces with explicit parametric control over all morphological attributes.

\paragraph*{Monocotyledonous Architecture (Maize)}

Monocot models implement a simplified two-level hierarchy:
\begin{enumerate}
    \item \textbf{Culm (main stem)}: Central axis with phyllotactic node arrangement (typically alternating 180$^\mathrm{o}$ azimuthal positioning). The culm extends beyond the terminal leaf by a configurable apical extension to represent the vegetative apex.
    \item \textbf{Leaf blades}: Linear-lanceolate leaves attached directly to nodes without intermediate branching structures (petioles/petiolules). Leaf attachment uses hard-graft geometry where the leaf base control points are sampled from the culm surface to ensure seamless integration.
\end{enumerate}
Maize leaves are characterized by linear geometry with monotonic width taper, longitudinal venation represented through camber curvature, and localized hinge bending at user-defined positions to capture characteristic blade drooping.

\paragraph{Hierarchical Parameter Inheritance.}
PD parameters follow a hierarchical precedence: global defaults specify baseline values for all organs; per-node overrides adjust parameters at specific nodes; and per-organ overrides (e.g., per-leaflet in trifoliates) provide fine-grained control. This inheritance structure minimizes parameter redundancy while preserving biological realism and user flexibility.

\paragraph{Species-Agnostic Generality.}
The architectural framework is species-agnostic (within monocot and dicot families): for any plant species, the same procedural pipeline applies. During hierarchical construction, the presence or absence of intermediate branching structures (petioles, petiolules) is determined by plant type. In particular:
\begin{itemize}
    \item Maize (monocot): Leaves attach directly to the stalk without branching structures. The algorithm proceeds from stalk construction directly to leaf attachment at nodes.
    \item Soybean/Mung Bean (dicot): Compound leaves require petiole branches from nodes, with petiolules branching from petioles and leaflets terminating petiolules.
\end{itemize}
This flexibility enables extension to other species (e.g., other grasses, legumes, or compound-leaved crops) by specifying the appropriate branching topology and morphological parameters in PDs without modifying the underlying procedural algorithms.

\subsection{LLM-Assisted Code Generation Process}
\label{sec:llm_process}
\begin{figure}[t!]
    \centering
    \includegraphics[trim=0cm 3cm 0cm 3cm, clip, width=0.8\linewidth]{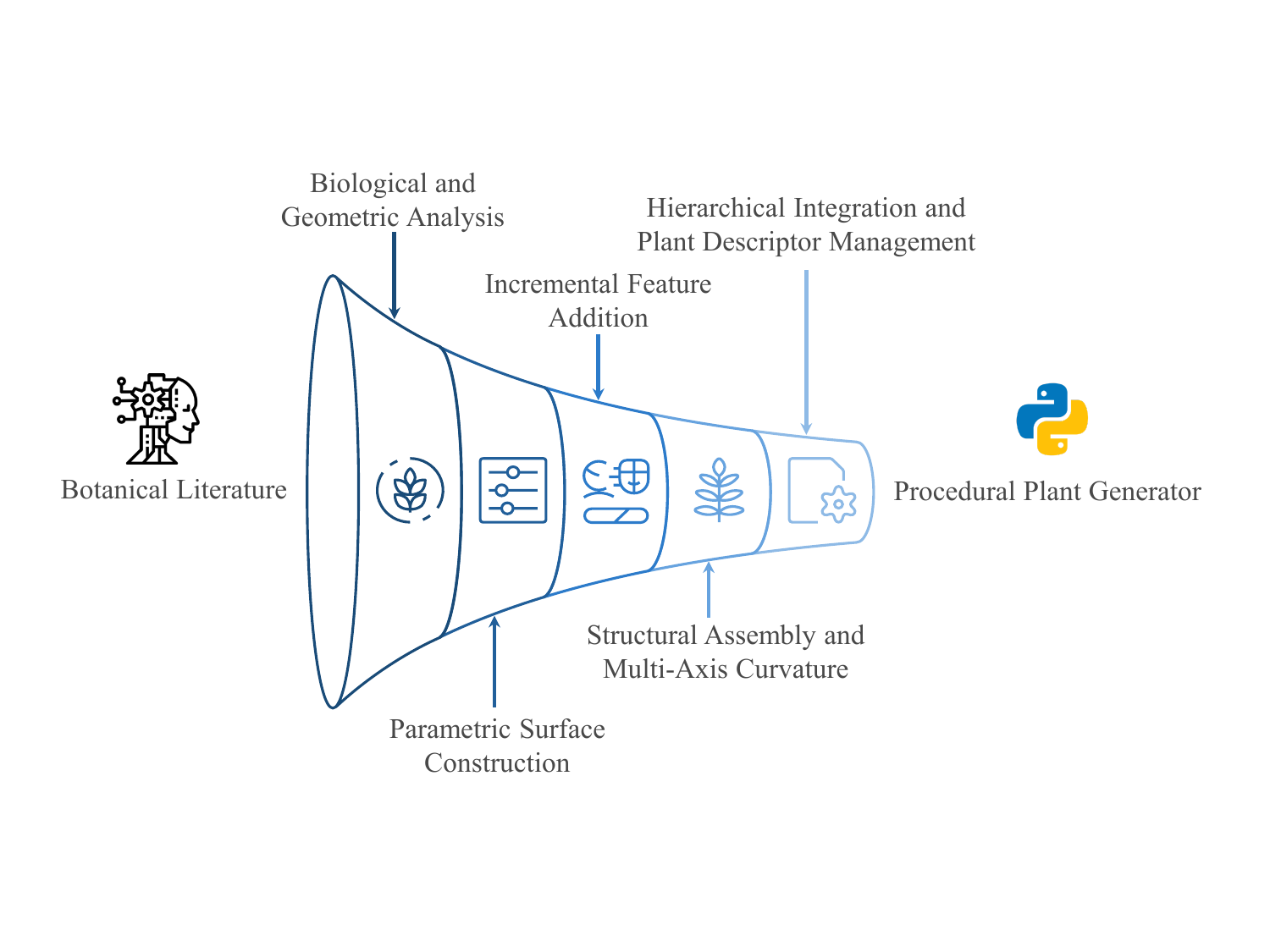}
    \caption{LLM-assisted iterative development workflow showing the five-stage progression from botanical literature analysis through parametric surface construction, incremental feature addition, structural assembly, to final hierarchical integration and PD management, resulting in PPG.}
    \label{fig:pipeline_llm}
\end{figure}

The modeling workflow originated from structured, iterative human–AI dialogues designed to refine domain-specific Python scripts progressively. Rather than attempting to generate complete solutions in a single step, we employed an iterative refinement strategy. Each conversation focused on solving a single, well-defined sub-problem before advancing to the next component. This approach mirrors how expert modelers incrementally build complex systems and has been shown to improve stability, interpretability, and convergence compared to monolithic generation approaches. \figref{fig:pipeline_llm} illustrates the five-stage iterative development workflow.

A state-of-the-art LLM was systematically guided through the following iterative development stages:

\paragraph*{Stage 1: Biological and Geometric Analysis.} 
The LLM was prompted to analyze botanical literature and generate detailed 3D geometric descriptions of target plant organs. This stage established the morphological vocabulary and geometric primitives required for subsequent modeling steps.

\paragraph*{Stage 2: Parametric Surface Construction}
Beginning with the smallest structural unit (the leaflet), the LLM generated initial Python code to construct B-spline surfaces using the \texttt{NURBS-Python (geomdl)} library \cite{bingol2019geomdl}. Early implementations produced geometric errors (e.g., inverted normals, collapsed control points, incorrect knot vectors), which were identified through visual inspection and fed back to the LLM with specific correction instructions. Reference images from botanical atlases, point cloud data from physical plants, and morphological constraints from literature (e.g., typical length-to-width ratios, curvature ranges) were progressively incorporated into the dialogue to constrain the solution space and improve biological realism.

For dicot leaflets, initial surfaces were flat and symmetric. Iterative refinement introduced asymmetric width profiles (controlled by separate left and right beta-function parameters), midrib offset, and localized apex/base shaping.

For monocot blades, the LLM initially implemented leaflets using dicot-style width profiles. Feedback specified the need for monotonic tapering to zero width at the tip and linear geometry, prompting the generation of the power-law width profile and tip-collapse control point arrangement.

\paragraph*{Stage 3: Incremental Feature Addition}
Once basic surface generation was validated, additional morphological realism was iteratively introduced through focused refinements, each tested and validated before proceeding:

\begin{itemize}[itemsep=0pt,topsep=0pt]
    \item \textbf{Leaf asymmetry:} Midrib offset, per-side width bias, and lateral skew parameters (see width profiles in \secref{sec:Mathematical_foundations}).
    \item \textbf{Global shape modifications:} Camber (cross-sectional bending; \eqnref{eq:camber}), longitudinal bend (gravitational droop for dicots, upward arch for monocots; \eqnref{eq:bend}), twist about the midrib (\eqnref{eq:twist}), and V-fold (dicot-specific folding; \eqnref{eq:vfold}).
    \item \textbf{Per-leaf orientation:} Yaw, pitch, and roll angles for leaflet attachment to petiolule/culm frames (frame construction in \secref{sec:Mathematical_foundations}).
    \item \textbf{Localized bending (monocot-specific):} Hinge-style transformations enabling sharp curvature at arbitrary positions along maize blades, with smoothstep-based transition zones to avoid control point discontinuities (\eqnref{eq:hinge}).
    \item \textbf{Hard-graft attachment (monocot-specific):} Replacement of leaf base control point rows with culm surface samples, including azimuthal arc sampling and radial offset control, to eliminate artificial gaps between blade and culm (\eqnref{eq:hardgraft}).
\end{itemize}

Each feature was added individually, with the LLM generating modified code, the user executing and visualizing outputs, and iterative corrections addressing geometric artifacts or parameter coupling issues (e.g., hinge bending causing unintended tip drift, which was resolved by rotating about per-column pivot points rather than a single global pivot).

\paragraph*{Stage 4: Structural Assembly and Multi-Axis Curvature}
After establishing leaf modules, the workflow proceeded to construct the stem, petiole, and petiolule using a node hierarchy and curved centerline generation. The LLM generated code implementing:

\begin{itemize}[itemsep=0pt,topsep=0pt]
    \item \textbf{Turtle-based centerline construction:} Inspired by the L-systems, we used the incremental tangent vector updates via rotation matrices, advancing position along curved trajectories.
    \item \textbf{Single-axis curvature:} Per-internode curvature \(\kappa\) (centerline curvature, \(\mathrm{m}^{-1}\); radius \(R{=}1/\kappa\); see \eqnref{eq:kappa}) and azimuth \(\phi\) defining the bending plane.
    \item \textbf{Multi-axis piecewise curvature:} Extension to support multiple simultaneous curvature terms with localized spans (e.g., spiral curvature via overlapping orthogonal bending planes), enabling complex 3D trajectories.
    \item \textbf{Parallel transport frames:} Twist-minimizing frame propagation for circular sweep surface construction.
    \item \textbf{Branch attachment frames:} Computation of local coordinate systems using Frenet frames  (tangent-normal-binormal) at petiole and petiolule base points using azimuth, pitch, and roll angles.
\end{itemize}

Errors in coordinate system conventions (e.g., azimuth mapping to incorrect bending directions), parameter broadcasting (e.g., a single curvature value applied per node vs. per petiole), and rotation composition order were identified through trial runs and corrected through iterative dialogue refinement.

\paragraph*{Stage 5: Hierarchical Integration and PD Management}
The final integration phase connected leaf modules to branch attachment points (petiolules for dicots, culm nodes for monocots) with per-node, per-petiole, and per-leaf parameter control. The LLM consolidated all command-line arguments into hierarchical PDs. This hierarchical parameter resolution enables complete specification of plant architecture state through structured, version-controllable data files, eliminating hard-coded parameters and enabling reproducibility across research groups.

\paragraph*{Pipeline Outputs}

The procedural pipeline produces three output formats, each optimized for distinct use cases:

\begin{itemize}[topsep=0pt,itemsep=0pt]
    \item \textbf{STL mesh files (.stl):} Tessellated triangular meshes derived from B-spline surface evaluation, suitable for visualization, rendering, and 3D printing. Tessellation density is controlled via surface sampling parameters (\texttt{sample\_size\_u}, \texttt{sample\_size\_v}) to balance geometric fidelity and file size. Note that the B-splines can also be adaptively sampled, generating more and smaller triangles in areas with higher curvature.
    
    \item \textbf{Analytic SMESH files (.dat):} Parametric surface metadata encoding B-spline control points, knot vectors, and degrees for all plant organs. SMESH format preserves the continuous mathematical representation, enabling direct import into isogeometric analysis frameworks, CAD software, and custom analysis pipelines without polygonal approximation.
    
    \item \textbf{YAML PD files (.yaml):} Human-readable parameter files recording all architectural and morphological parameters, ensuring complete reproducibility of generated models. PDs serve as version-controllable ``plant recipes'', enabling collaborative model refinement and systematic parameter space exploration.
\end{itemize}

\section{Mathematical Foundations}
\label{sec:Mathematical_foundations}

Plant organs in FloraForge are represented as tensor product B-spline surfaces, providing smooth, mathematically continuous geometric primitives essential for analysis-ready modeling. B-splines are piecewise polynomial curves and surfaces constructed as linear combinations of basis functions, widely used in computer-aided design (CAD)  for their flexibility and mathematical elegance. This section establishes the mathematical formulations underlying the procedural generation algorithms.

\subsection{B-Spline Basis Functions}

At the core of B-splines are basis functions defined over a knot vector. A B-spline curve of degree \(p\) is defined by a set of \(n+1\) control points \(\{\mathbf{P}_i\}_{i=0}^{n}\) and a knot vector:
\begin{equation}
\mathbf{U} = \{u_0, u_1, \ldots, u_{n+p+1}\},
\end{equation}
containing non-decreasing real numbers that partition the parametric space. The B-spline basis functions \(N_{i,p}(u)\) are defined recursively using the Cox-de Boor formula. Starting with piecewise constant functions for \(p=0\):
\begin{equation}
N_{i,0}(u) = 
\begin{cases}
1 & \text{if } u_i \leq u < u_{i+1} \\
0 & \text{otherwise}.
\end{cases}
\end{equation}
For higher degrees \(p = 1, 2, 3, \ldots\), the basis functions are computed recursively:
\begin{equation}
N_{i,p}(u) = \frac{u - u_i}{u_{i+p} - u_i} N_{i,p-1}(u) + \frac{u_{i+p+1} - u}{u_{i+p+1} - u_{i+1}} N_{i+1,p-1}(u).
\label{eqn:basis}
\end{equation}
The curve is then expressed parametrically as a weighted sum of control points:
\begin{equation}
\mathbf{C}(u) = \sum_{i=0}^{n} N_{i,p}(u) \, \mathbf{P}_i, \quad u \in [u_p, u_{n+1}].
\end{equation}
These basis functions possess several key properties that make them suitable for plant modeling: \textit{local support} (each \(N_{i,p}(u)\) is non-zero only over \(p+1\) knot spans, enabling local shape control), \textit{partition of unity} (\(\sum_{i=0}^{n} N_{i,p}(u) = 1\) for all~\(u\), ensuring affine invariance), and \textit{non-negativity} (\(N_{i,p}(u) \geq 0\)), ensuring smooth interpolation and intuitive geometric control.

\subsection{B-Spline Surface Formulation}

B-spline surfaces extend the concept of B-spline curves into two dimensions using tensor products. Given two knot vectors \(\mathbf{U} = \{u_0, \ldots, u_{n+p+1}\}\) and \(\mathbf{V} = \{v_0, \ldots, v_{m+q+1}\}\), along with an \((n+1) \times (m+1)\) grid of control points \(\mathbf{P}_{i,j}\), the B-spline surface is defined as:
\begin{equation}
\mathbf{S}(u,v) = \sum_{i=0}^{n} \sum_{j=0}^{m} N_{i,p}(u) \, N_{j,q}(v) \, \mathbf{P}_{i,j}, \quad 0 \leq u, v \leq 1
\label{B-spline_surface}
\end{equation}
where \(N_{i,p}(u)\) and \(N_{j,q}(v)\) are one-dimensional B-spline basis functions (\eqnref{eqn:basis}) corresponding to the respective knot vectors, with degrees \(p\) and \(q\) in the \(u\) and \(v\) parametric directions, respectively. This tensor product structure enables independent control of surface smoothness and shape in both parametric directions.

Non-Uniform Rational B-Splines (NURBS) extend B-splines by introducing weights \(w_{i,j}\) for each control point, enabling exact representation of conic sections (circles, ellipses) and improved control over surface shape.
However, in this work, all plant organ surfaces are represented as non-rational B-splines with uniform weights (\(w_{i,j} = 1\) for all control points), reducing to \eqnref{B-spline_surface}. This choice is motivated by three considerations. First, plant organs (leaves, stems, branches) do not require exact conic section representation; the polynomial basis of B-splines provides sufficient geometric complexity to capture organic shapes with high biological fidelity. Second, non-rational formulations eliminate weight normalization computations, reducing computational overhead during surface evaluation and tessellation while maintaining numerical stability. Third, without weight parameters, the relationship between control point positions and surface geometry is more intuitive, simplifying manual parameter adjustment and automated fitting procedures. The resulting B-spline representation provides smooth, mathematically continuous surfaces with explicit parametric control, enabling direct editing through PD parameters and compatibility with isogeometric analysis workflows.

\subsection{Rotation and Frame Transport}

Procedural construction of curved stems and branches requires smooth propagation of local coordinate frames (tangent \(\mathbf{T}\), normal \(\mathbf{N}\), binormal \(\mathbf{B}\)) along centerline curves. Rotation about an arbitrary axis \(\mathbf{a}\) by angle \(\theta\) is computed using Rodrigues' rotation formula:
\[
\mathbf{R}(\mathbf{a}, \theta) = \mathbf{I} + \sin(\theta) \, [\mathbf{a}]_{\times} + (1 - \cos(\theta)) \, [\mathbf{a}]_{\times}^2
\]
where \([\mathbf{a}]_{\times}\) is the skew-symmetric cross-product matrix of the unit axis vector \(\mathbf{a}\):
\[
[\mathbf{a}]_{\times} = 
\begin{bmatrix}
0 & -a_z & a_y \\
a_z & 0 & -a_x \\
-a_y & a_x & 0
\end{bmatrix}.
\]
Expanding Rodrigues' formula explicitly:
\[
\mathbf{R}(\mathbf{a}, \theta) = 
\begin{bmatrix}
a_x^2 C + c & a_x a_y C - a_z s & a_x a_z C + a_y s \\
a_y a_x C + a_z s & a_y^2 C + c & a_y a_z C - a_x s \\
a_z a_x C - a_y s & a_z a_y C + a_x s & a_z^2 C + c
\end{bmatrix},
\]
where \(c = \cos(\theta)\), \(s = \sin(\theta)\), and \(C = 1 - \cos(\theta)\).

Parallel transport frames are computed along the stem centerline polyline \(\{\mathbf{p}_0, \mathbf{p}_1, \ldots, \mathbf{p}_M\}\) to minimize frame twist. At each vertex \(i\), the tangent vector \(\mathbf{T}_i\) is computed from adjacent points, and the normal \(\mathbf{N}_i\) is propagated by rotating the previous normal \(\mathbf{N}_{i-1}\) into the plane perpendicular to \(\mathbf{T}_i\):
\[
\mathbf{N}_i = \mathbf{R}(\mathbf{a}_i, \alpha_i) \, \mathbf{N}_{i-1}, \quad \mathbf{a}_i = \frac{\mathbf{T}_{i-1} \times \mathbf{T}_i}{\|\mathbf{T}_{i-1} \times \mathbf{T}_i\|}, \quad \alpha_i = \arccos(\mathbf{T}_{i-1} \cdot \mathbf{T}_i)
\]
The binormal vector is obtained as \(\mathbf{B}_i = \mathbf{T}_i \times \mathbf{N}_i\). This parallel transport procedure ensures smooth, twist-free orientation frames for the generalized cylinder construction.

\subsection{Stem and Branch Curvature via Piecewise Multi-Axis Bending}

Stem and branch curvature is controlled through curvature parameters \(\kappa\) applied incrementally along discretized segments. 
\begin{equation}
\kappa(s) = \left\lVert \frac{d\mathbf{T}}{ds} \right\rVert, \qquad
\mathbf{T}(s)=\frac{d\mathbf{C}}{ds}, \qquad
R(s)=\frac{1}{\kappa(s)}.
\label{eq:kappa}
\end{equation}

For each internode segment of length \(L\), divided into \(n_{\text{seg}}\) steps with stepsize \(ds = L / n_{\text{seg}}\), the tangent direction \(\mathbf{t}\) is updated via rotation about a specified axis \(\mathbf{a}\):
\[
\mathbf{t}_{k+1} = \mathbf{R}(\mathbf{a}, \kappa \cdot ds) \, \mathbf{t}_k.
\]
The next centerline point is then computed as:
\[
\mathbf{p}_{k+1} = \mathbf{p}_k + \mathbf{t}_{k+1} \, ds.
\]
For single-axis curvature, the bending plane normal (rotation axis) is defined by the azimuth angle \(\phi\) in the global \(xy\)-plane:
\[
\mathbf{a}(\phi) = \begin{bmatrix} -\sin(\phi) \\ \cos(\phi) \\ 0 \end{bmatrix}.
\]
This parameterization maps azimuth \(\phi = 0^\circ\) to the \(+y\) axis and \(\phi = 90^\circ\) to the \(-x\) axis, enabling intuitive directional control of stem bending.

For multi-axis piecewise curvature (implemented via \texttt{*-kappa-terms} in PDs), multiple curvature terms \(\{(\kappa_j, \mathbf{a}_j, [s_0^j, s_1^j])\}\) can be applied simultaneously within specified fractional spans \([s_0, s_1] \subset [0,1]\) of an internode. At each step fraction \(s = k / n_{\text{seg}}\), all active terms (where \(s_0^j \leq s \leq s_1^j\)) contribute rotations sequentially:
\[
\mathbf{t}_{k+1} = \prod_{j \in \text{active}(s)} \mathbf{R}(\mathbf{a}_j, \kappa_j \cdot ds) \, \mathbf{t}_k.
\]
This formulation enables localized bending, spiral curvature, and complex 3D stem trajectories through the addition of curvature components.

\subsection{Circular Sweep Surface Construction}

Stems, petioles, and petiolules are represented as generalized cylinder B-spline surfaces generated by sweeping circular cross-sections along curved centerline polylines. Given a centerline \(\{\mathbf{p}_i\}_{i=0}^{M}\), radius profile \(\{r_i\}_{i=0}^{M}\), and parallel transport frames \(\{(\mathbf{T}_i, \mathbf{N}_i, \mathbf{B}_i)\}_{i=0}^{M}\), the control point grid is constructed by sampling \(U\) points around circular rings at each centerline vertex:
\[
\mathbf{P}_{u,v} = \mathbf{p}_v + r_v \left( \cos(\theta_u) \, \mathbf{N}_v + \sin(\theta_u) \, \mathbf{B}_v \right)
\]
where \(\theta_u = 2\pi u / U\) for \(u \in \{0, 1, \ldots, U\}\) (with \(u=U\) duplicating \(u=0\) to close the surface seam) and \(v \in \{0, 1, \ldots, M\}\). The resulting control point grid of size \((U+1) \times (M+1)\) defines a B-spline surface with degrees \(p_u\) and \(p_v\) (typically 2 and 3, respectively) in the circumferential and axial directions.

\subsection{Leaf Surface Geometry: Parametric Width Profiles and Deformations}

Leaf blades (both dicot and monocot blades) are constructed as B-spline surfaces defined by the control point matrices \(\mathbf{P}_{i,j}\) with dimensions \(M_u \times M_v\) (longitudinal \(\times\) transverse). The base control matrix is generated in a local coordinate system with the \(x\)-axis aligned with the leaf midrib and the \(yz\)-plane spanning the leaf width.

\paragraph{Width Profile Formulation.}
For monocot blades (maize), the geometry is a ruled surface where the leaf width tapers monotonically from base width \(W_0\) to zero at the tip using a power-law profile:
\begin{equation}
W(u) = W_0 \, (1 - u^p), \quad u \in [0, 1],
\label{eq:width}
\end{equation}
where \(u\) is the normalized longitudinal coordinate (\(u=0\) at base, \(u=1\) at tip) and \(p\) controls the taper rate (default \(p \approx 1.2\)).

For dicot leaflets (soybean, mung bean), asymmetric width profiles are generated using a beta-function-inspired formulation:
\[
W(u; \alpha, \beta) = W_{\max} \, u^\alpha \, (1-u)^\beta
\]
normalized such that \(\max_u W(u) = W_{\max}\). Parameters \(\alpha\) and \(\beta\) control the position and sharpness of maximum width, enabling the representation of diverse leaf shapes (ovate, lanceolate, and linear). Additional localized modulation functions sharpen the apex and round the base:
\[
W_{\text{final}}(u) = W(u) \cdot \left(1 - 0.35 \, a_{\text{apex}} \, f_{\text{tip}}(u)\right) \cdot \left(1 + 0.20 \, a_{\text{base}} \, f_{\text{base}}(u)\right),
\]
where constants 0.35 and 0.20 are empirically determined shape coefficients, \(f_{\text{tip}}(u) = \max(0, (u-0.8)/0.2)\) and \(f_{\text{base}}(u) = \max(0, (0.15-u)/0.15)\) are ramp functions, and \(a_{\text{apex}}, a_{\text{base}}\) are shape-specific coefficients.

\paragraph{Leaf Centerline and Longitudinal Curvature.}
The leaf midrib centerline \(\mathbf{C}(u)\) in local coordinates is constructed with optional lateral skew and longitudinal bending:
\[
\mathbf{C}(u) = \begin{bmatrix} L \, u \\ y_{\text{skew}}(u) \\ z_{\text{bend}}(u) \end{bmatrix},
\]
where \(L\) is leaf length, \(y_{\text{skew}}(u) = \alpha_{\text{skew}} W_0 \sin(\pi u)\) introduces lateral asymmetry, and \(z_{\text{bend}}(u) = \beta_{\text{bend}} L \sin(\pi u) \label{eq:bend}\) creates an upward (or downward) arch. For dicots, the sign convention is negative (\(\beta_{\text{bend}} < 0\)) to produce gravitational droop; for monocots, positive values produce upward curvature typical of grass leaves.

\paragraph{Cross-Sectional Camber.}
Transverse curvature (camber) is applied perpendicular to the midrib using a longitudinally varying amplitude:
\[
z_{\text{camber}}(u, v) = \alpha_{\text{camber}} L \, (1 - |v|)^{p_{\text{camber}}} \, \sin(\pi u),
\label{eq:camber}
\]
where \(v \in [-1, 1]\) is the transverse coordinate (negative = left, positive = right), \(\alpha_{\text{camber}}\) is the camber amplitude coefficient, and \(p_{\text{camber}}\) controls the decay rate from midrib to edge (higher values produce flatter edges).

\paragraph{Twist and V-Fold.}
Longitudinal twist is introduced by rotating the local transverse frame progressively along the midrib:
\[
\theta_{\text{twist}}(u) = \alpha_{\text{twist}} \, u
\label{eq:twist}
\]
applied via rotation matrix \(\mathbf{R}(\mathbf{T}(u), \theta_{\text{twist}}(u))\) where \(\mathbf{T}(u)\) is the local tangent to the midrib.

V-folding along the midrib (common in dicot leaflets) is implemented by rotating control points about the tangent vector \(\mathbf{T}(u)\) with an angle proportional to transverse position:
\[
\theta_{\text{fold}}(u, v) = \alpha_{\text{fold}} \, \text{sign}(v) \, |v|^{p_{\text{fold}}} \, \left[\sin(\pi u)\right]^{q_{\text{fold}}},
\label{eq:vfold}
\]
where \(p_{\text{fold}}\) controls the fold sharpness and \(q_{\text{fold}}\) modulates the envelope along the leaf length.

\paragraph{Localized Hinge Bending (Monocot).}
Maize leaves often exhibit sharp localized bends at specific positions along the blade. This is implemented via hinge transformations applied to control point rows distal to a specified normalized position \(u_0\):
\[
\mathbf{P}_{i,j}^{\text{hinge}} = \mathbf{P}_{i_0, j} + \mathbf{R}(\mathbf{a}_{\text{hinge}}, w_i \, \theta_{\text{hinge}}) \, (\mathbf{P}_{i,j} - \mathbf{P}_{i_0, j}),
\label{eq:hinge}
\]
where \(i_0\) corresponds to \(u_0\), \(\mathbf{a}_{\text{hinge}}\) is the hinge rotation axis (typically the binormal \(\mathbf{B}_{i_0}\)), \(\theta_{\text{hinge}}\) is the hinge angle, and \(w_i\) is a smooth ramp weight:
\[
w_i = \begin{cases}
0 & u_i < u_0 \\
\text{smoothstep}\left(\frac{u_i - u_0}{\sigma}\right) & u_i \geq u_0
\end{cases}, \quad \text{smoothstep}(x) = x^2(3 - 2x)
\]
with transition width \(\sigma\). Multiple hinges can be applied sequentially to create complex sigmoidal curvature.

\begin{algorithm}[t!]
    \caption{Unified Procedural Plant Generation (Monocot \& Dicot)}
    \label{alg:unified}
    \footnotesize
\KwIn{YAML PD: node-z, stem geometry, leaf parameters}
\KwOut{STL mesh and SMESH parametric surface files}

\tcp{Stage 1: Build stem/culm centerline via turtle-based integration}
Load YAML: node heights \(\{z_i\}\), diameters \(\{d_i\}\), curvature \(\{\kappa_i\}\)\;
Initialize: \(\mathbf{p} \leftarrow [0,0,0]\), \(\mathbf{t} \leftarrow [0,0,1]\)\;
\For{each internode $i$}{
    \For{integration steps}{
        Apply curvature: \(\mathbf{t} \leftarrow \mathbf{R}(\mathbf{a}_i, \kappa_i \cdot ds) \cdot \mathbf{t}\)\;
        Advance: \(\mathbf{p} \leftarrow \mathbf{p} + \mathbf{t} \cdot ds\)\;
    }
}
\If{monocot}{Extend culm beyond terminal leaf\;}
Compute parallel transport frames \(\{(\mathbf{T}_j, \mathbf{N}_j, \mathbf{B}_j)\}\)\;
Generate stem surface via circular sweep\;

\tcp{Stage 2: Branch and leaf construction}
\For{each node $i$}{
    \If{dicot}{
        \For{each petiole at node $i$}{
            Build petiole centerline with curvature\;
            Generate petiole B-spline surface\;
            \For{trifoliate sides: Terminal, Left, Right}{
                Build petiolule centerline\;
                Generate petiolule B-spline surface\;
                \If{leaflet enabled}{
                    Generate leaflet control grid (beta-function width, camber, V-fold)\;
                    Orient to petiolule tip frame\;
                    Create leaflet B-spline surface\;
                }
            }
        }
    }
    
    \If{monocot}{
        \If{leaf enabled at node $i$}{
            Generate blade control grid (power-law taper, camber, bend)\;
            Apply hinge bends at specified positions\;
            \tcp{Hard-graft: replace base rows with culm surface samples}
            \For{graft rows $g = 0, \ldots, n_{\text{graft}}-1$}{
                Sample culm surface points at node height\;
                Replace control grid base with culm samples\;
            }
            Create blade B-spline surface\;
        }
    }
}

\tcp{Stage 3: Export}
Tessellate all B-spline surfaces \(\rightarrow\) STL mesh\;
Export B-spline metadata \(\rightarrow\) SMESH files\;
\end{algorithm}

\paragraph{Hard-Graft Leaf Base (Monocot).}
For realistic attachment of maize leaves to the culm, the first \(n_{\text{graft}}\) rows of the leaf control point grid are replaced with points sampled from the culm surface. At each graft row \(g \in \{0, 1, \ldots, n_{\text{graft}}-1\}\), positioned at height:
\[
z_g = z_{\text{node}} + g \cdot \frac{\Delta z_{\text{axial}}}{n_{\text{graft}} - 1},
\]
the control points are sampled from a circular arc on the culm surface centered at azimuth \(\phi_{\text{leaf}}\) with arc span \(\Delta\phi\):
\[
\mathbf{P}_{g,j} = \mathbf{p}_{\text{culm}}(z_g) + (r_{\text{culm}}(z_g) + \delta_r) \left[ \cos(\phi_j) \, \mathbf{N}_{\text{culm}}(z_g) + \sin(\phi_j) \, \mathbf{B}_{\text{culm}}(z_g) \right],
\label{eq:hardgraft}
\]
where \(\phi_j = \phi_{\text{leaf}} + (2j/(M_v - 1) - 1) \, \Delta\phi/2\), \(r_{\text{culm}}(z_g)\) is the culm radius at height \(z_g\), \(\delta_r\) is a small radial offset, and \(\mathbf{N}_{\text{culm}}, \mathbf{B}_{\text{culm}}\) are the culm normal and binormal vectors. This grafting procedure ensures seamless integration of the leaf surface with the culm geometry, eliminating artificial gaps or penetrations.

An optional soft blending zone can be applied to rows immediately beyond the graft region (\(g > n_{\text{graft}}\), \(u < u_{\text{attach}}\)) using linear interpolation:
\[
\mathbf{P}_{i,j}^{\text{blend}} = (1 - \alpha_i) \, \mathbf{P}_{i,j}^{\text{leaf}} + \alpha_i \, \mathbf{P}_{i,j}^{\text{culm-ring}}, \quad \alpha_i = 1 - \text{smoothstep}\left(\frac{u_i}{u_{\text{attach}}}\right)
\]
ensuring a smooth transition from rigid graft to free blade geometry.

\subsection{Framework Implementation} 
\label{sec:implementation}

\begin{figure}[t!]
    \centering
    \includegraphics[trim=0cm 4.5cm 0cm 4.5cm, clip, width=0.85\linewidth]{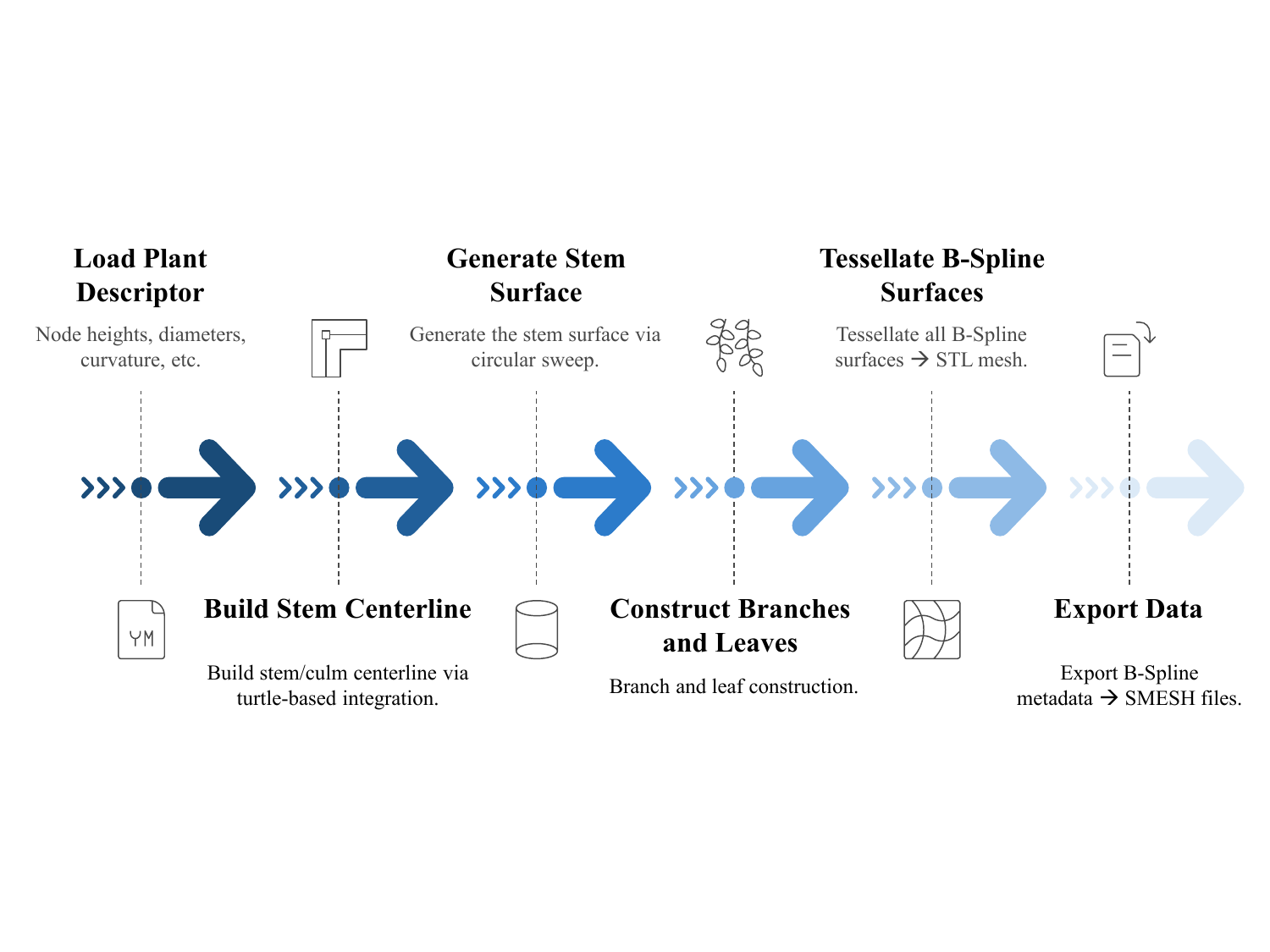}
    \caption{Procedural generation workflow showing the computational pipeline from PD as input through stem centerline construction, branch/leaf assembly, and dual-format export (STL mesh and SMESH parametric surfaces).}
    \label{fig:pipeline_script}
\end{figure}

Our method encapsulates the species-agnostic procedural generation workflow, with conditional branching for species-specific features. \figref{fig:pipeline_script} illustrates the computational workflow from PD loading through B-spline surface generation to final export. The complete algorithm (see \algref{alg:unified}) consists of 3 stages. Stage 1 constructs stem or culm centerlines via incremental turtle-based integration with Rodrigues rotation for curvature application, followed by parallel transport frame computation and circular sweep surface generation. Stage 2 branches based on botanical architecture: dicotyledonous species (soybean, mung bean) construct hierarchical compound leaf structures through nested petiole→petiolule→leaflet levels, generating 7 B-spline surfaces per trifoliate set (1 petiole + 3 petiolules + 3 leaflets); monocotyledonous species (maize) implement simplified architecture with leaf blades attaching directly to culm nodes via hard-graft integration, where basal \(n_{\text{graft}}\) control point rows are sampled from the culm surface to eliminate attachment gaps, generating 1 surface per node. Stage 3 performs species-agnostic tessellation and dual-format export (STL for visualization, SMESH for parametric representation). The framework enables extension to arbitrary plant species through PD without algorithmic modification.

\paragraph{PD Structure and Parametric Editability}

The human-readable PD format enables domain experts to specify 
complex plant architectures without programming knowledge. \figref{fig:yaml_configs} 
demonstrates the hierarchical parameter structure for two morphologically distinct 
maize plants, highlighting how architectural variation is achieved through 
parameter editing.

\section{Results}
\label{sec:results}

The procedural modeling pipeline successfully generated biologically accurate, analysis-ready plant models for multiple species and developmental stages. We demonstrate this capability across 8 plant instances (5 maize genotypes and 3 mung bean plants), achieving sub-centimeter geometric fidelity with mean symmetric Chamfer distances of 0.021\,m for maize and 0.005\,m for mung bean when validated against empirical point cloud data (Sections~\ref{subsec:maize_models} and~\ref{subsec:mungbean_models}).

\subsection{Dataset}
\label{subsec:dp}

The maize data used in this study were selected from the publicly available \textit{MaizeField3D} dataset~\citep{maizefield3d}, which provides multi-genotype, field-grown maize point clouds acquired using LiDAR-based scanning at Iowa State University. Five plants (T8, CML69, M162W, CML238, and CI90C) were chosen from this dataset, corresponding to genotypes representative of diverse morphological architectures. Each point cloud dataset was used to manually fit the procedural models described below.

The mung bean data used in this study were collected from experimental fields at Iowa State University’s Agricultural Engineering and Agronomy Research Farm. Each genotype was grown in a separate 5-foot, three-row plot, containing approximately 100 plants per plot. Plants were sampled at the R3 growth stage. 3D models of mung bean plants were generated from iPhone video recordings using a Neural Radiance Field (NeRF)-based 3D reconstruction method~\citep{mildenhall2021nerf}. The resulting 3D point clouds from three randomly selected samples, each representing a different genotype, were used for procedural model generation.


\subsection{Procedurally Generated Plant Models}
We demonstrate the pipeline's capability to generate editable, analysis-ready model instances by fitting procedural templates to empirical point cloud data for multiple plant species and developmental stages.

\subsubsection{Maize Models}
\label{subsec:maize_models}

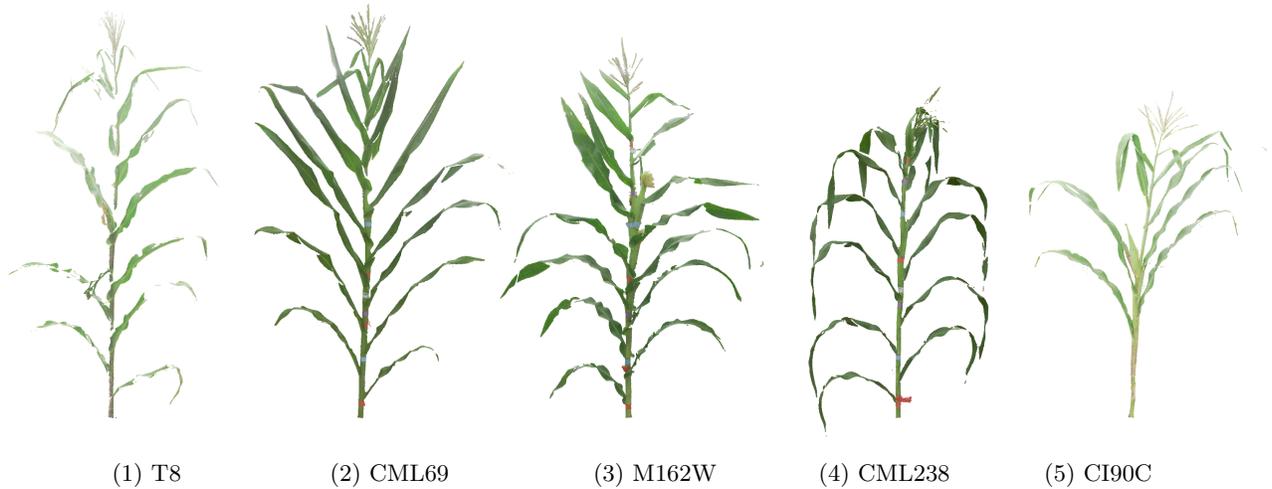
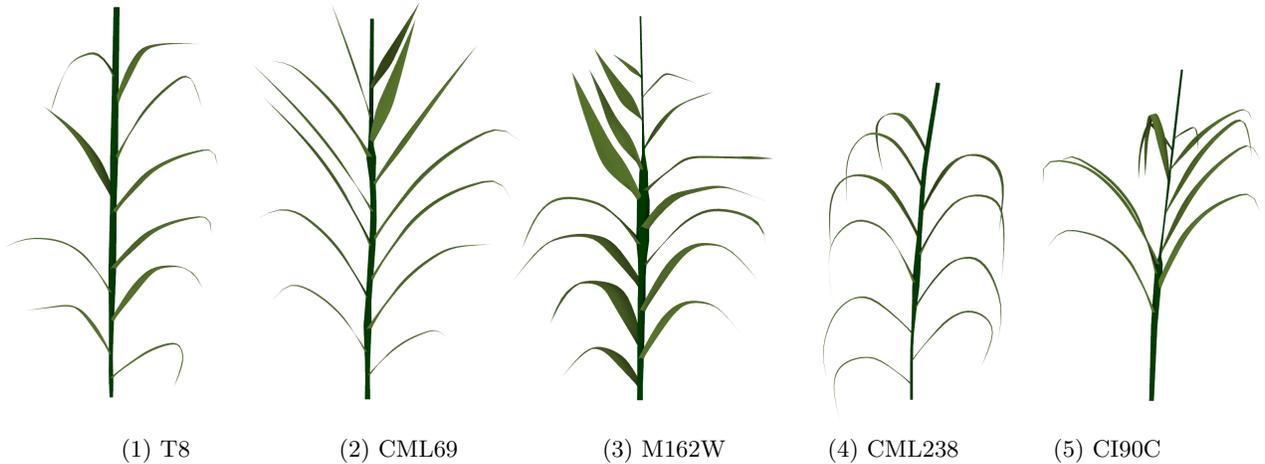
\begin{figure}[t!]
    \centering
    \begin{subfigure}{\textwidth}
        \centering
        \includegraphics[trim=0cm 20cm 0cm 20cm, clip, width=\textwidth]{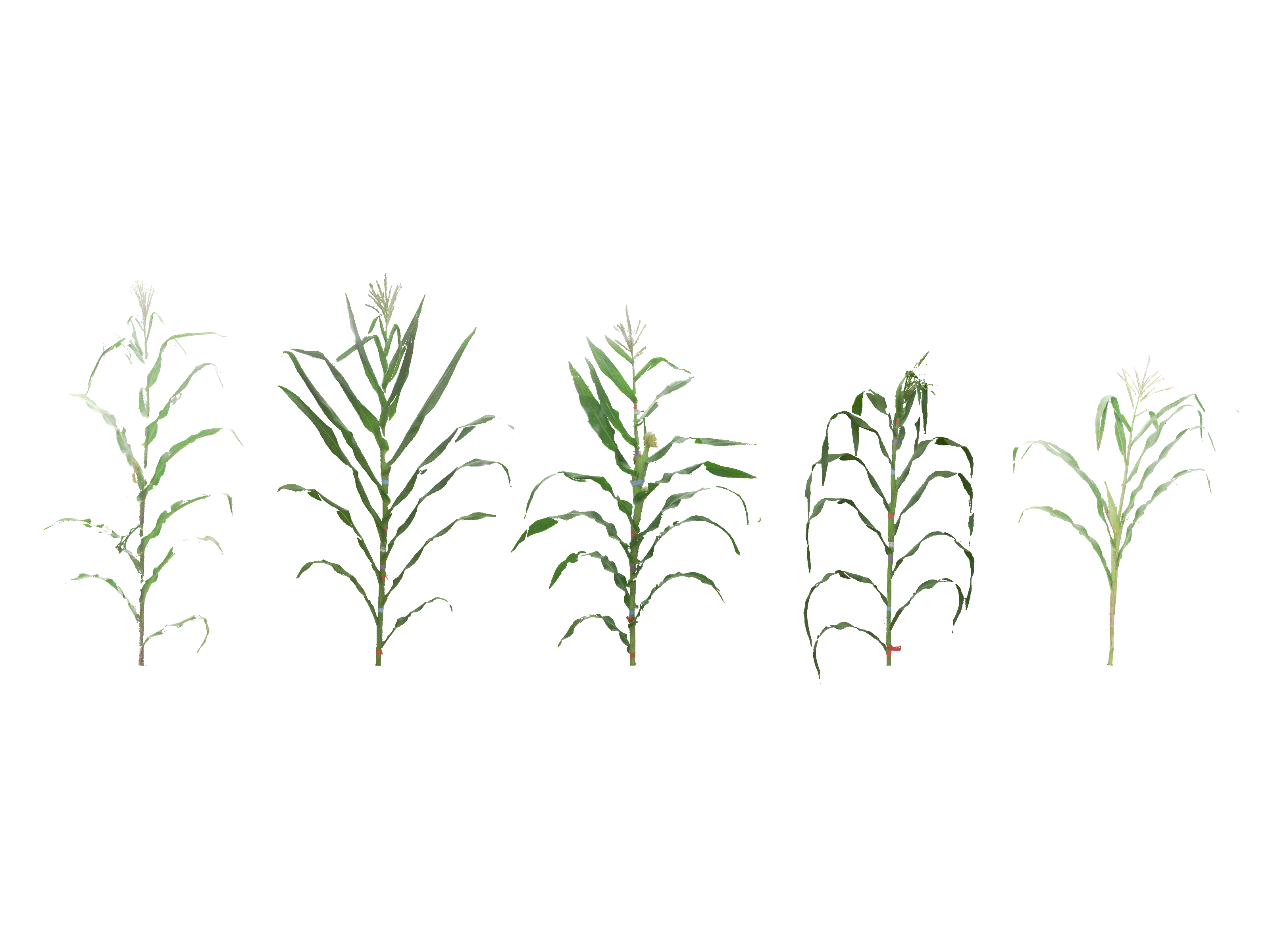}
        \setlength{\unitlength}{1cm}
        \begin{picture}(0,0)
            \put(-6.9,0.5){\footnotesize (1) T8}
            \put(-4.0,0.5){\footnotesize (2) CML69}
            \put(-0.5,0.5){\footnotesize (3) M162W}
            \put(2.5,0.5){\footnotesize (4) CML238}
            \put(5.5,0.5){\footnotesize (5) CI90C}
        \end{picture}
        \caption{Input point cloud data of real plants from LiDAR scanning.}
        \label{fig:maize_original}
    \end{subfigure}
    \begin{subfigure}{\textwidth}
        \centering
        \includegraphics[trim=-3cm 22cm -4cm 20cm, clip, width=\textwidth]{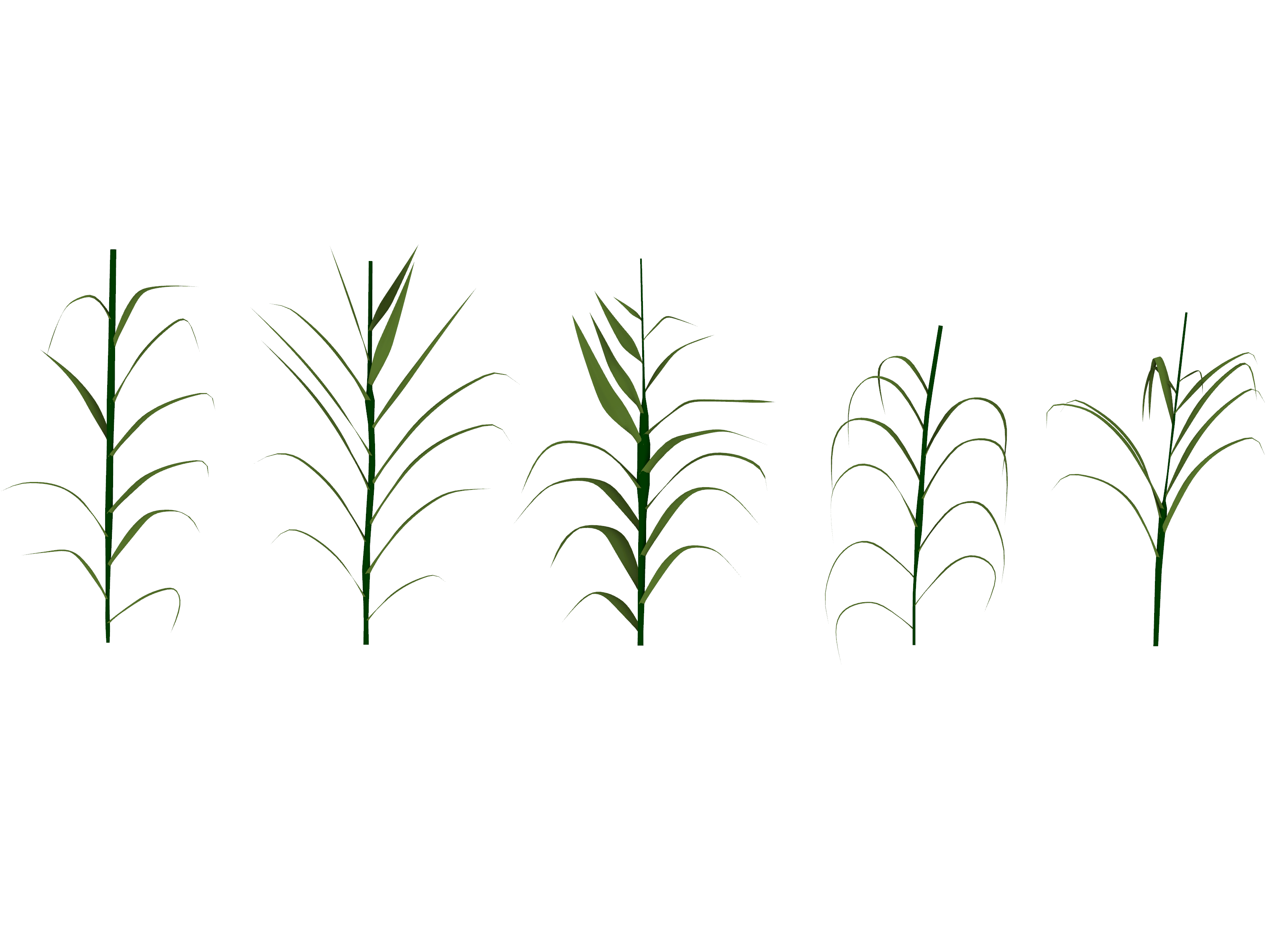}
        \setlength{\unitlength}{1cm}
        \begin{picture}(0,0)
            \put(-6.9,0.5){\footnotesize (1) T8}
            \put(-4.0,0.5){\footnotesize (2) CML69}
            \put(-0.5,0.5){\footnotesize (3) M162W}
            \put(2.5,0.5){\footnotesize (4) CML238}
            \put(5.5,0.5){\footnotesize (5) CI90C}
        \end{picture}
        \caption{Corresponding procedurally generated models.}
        \label{fig:maize_procedural}
    \end{subfigure}
    \caption{Comparison of empirical LiDAR point cloud data (a) and fitted procedural models (b) for five maize genotypes. The procedural models accurately reproduce plant-specific architectural features, including culm trajectories, leaf orientations, and blade morphology, while maintaining continuous B-spline surface representations suitable for downstream computational analysis.}
    \label{fig:maize_fitted}
\end{figure}

Five maize plants representing diverse genotypes were generated by iteratively adjusting PD to achieve geometric correspondence with LiDAR point cloud reconstructions. \figref{fig:maize_fitted} compares the scanned point clouds with the LLM-assisted generated 3D model instances, demonstrating the accurate capture of key morphological features: culm curvature and diameter variation, distichous phyllotactic leaf arrangement, hard-graft leaf attachment geometry, and complex blade morphology, including localized hinge bending and monotonic width tapering. Each model required approximately 2-6 hours of parameter adjustment to achieve geometric correspondence with the point cloud data. The PDs for these five plants demonstrate the range of morphological variation achievable through parameter editing.

We quantify geometric fidelity of the fitting using the symmetric Chamfer distance (CD). Across five maize genotypes, manual tuning of the LLM-seeded template reduces CD from \textbf{0.1\,m} on average to \textbf{0.02\,m} (mean reduction $\approx$\textbf{79\%}; \figref{fig:cd_maize_template_instance}). Per-genotype reductions range from \textbf{72\%} to \textbf{85\%} (3.6--6.8$\times$ lower CD). This indicates that while the LLM-seeded template provides a plausible initial geometry, targeted adjustments (node heights, leaf length/width, pitch, and localized hinge bends) substantially improve agreement with the point cloud.

To make the starting point explicit, we visualize an LLM-seeded
\emph{initial template} behind each fitted maize model in \figref{fig:maize_baseline}.
For every genotype, the initial template PD is drafted by the LLM from the
approximate node heights and the observed number of leaves, with leaves
placed using a default 0°/180° distichous phyllotaxy and no manual tuning (e.g., no hinge bends, no node-specific curvature overrides). The foreground shows the manually tuned instance. The visual gap between the two aligns with the CD reductions reported in \figref{fig:cd_maize_template_instance}.

\begin{figure}[!t]
  \centering
  \includegraphics[trim=0cm 22cm 0cm 25cm, clip, width=\linewidth]{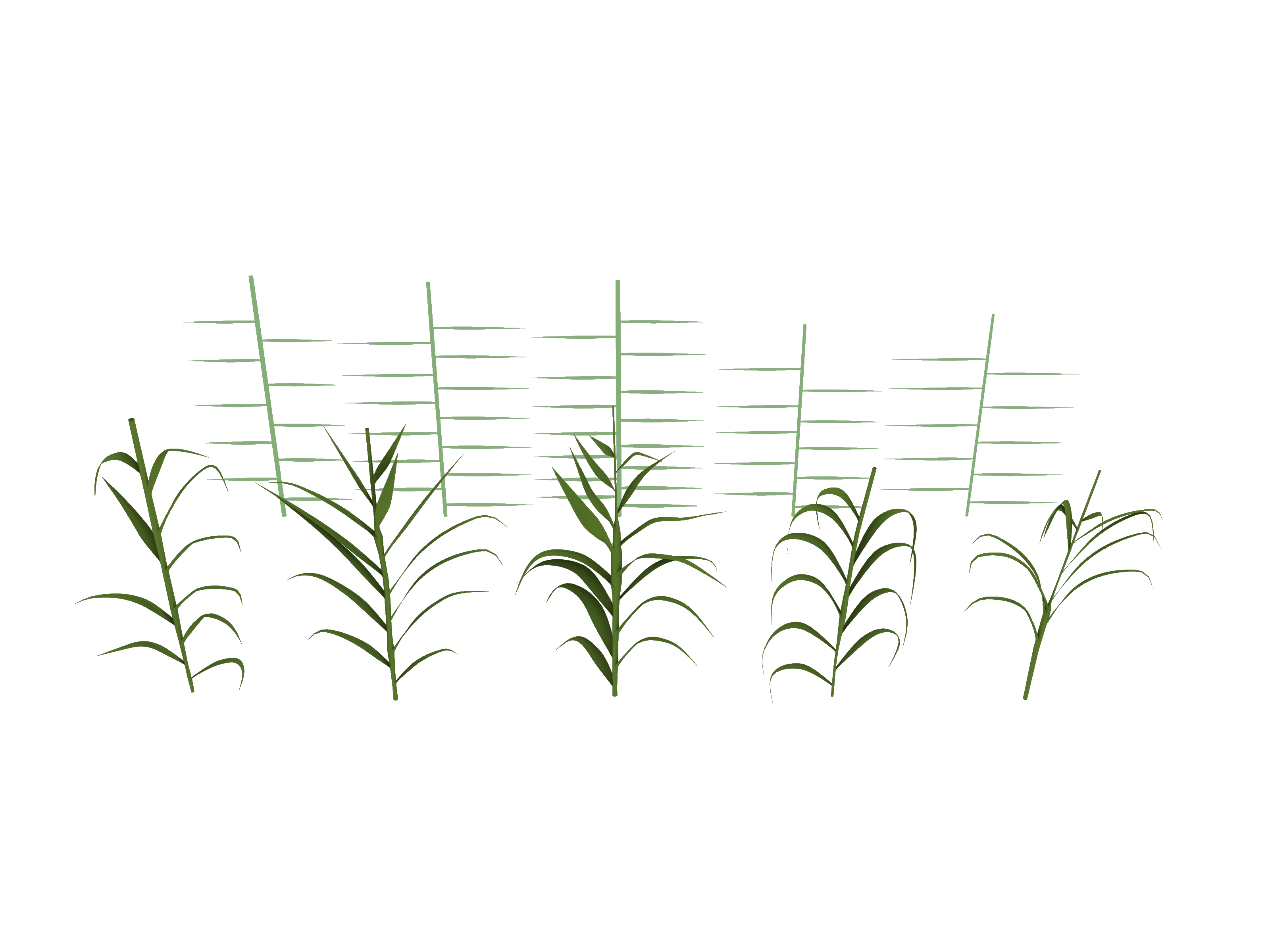}
    \setlength{\unitlength}{1cm}
    \begin{picture}(0,0)
        \put(-6.2,0){\footnotesize (1) T8}
        \put(-3.9,0){\footnotesize (2) CML69}
        \put(-1.0,0){\footnotesize (3) M162W}
        \put(1.7,0){\footnotesize (4) CML238}
        \put(4.5,0){\footnotesize (5) CI90C}
    \end{picture}
  \caption{Foreground: manually tuned procedural model instance.
  Background: LLM-seeded \emph{initial template} generated automatically from
  (i) approximate node $z$-positions and (ii) the observed number of leaves for each plant; leaves are placed using a default distichous phyllotaxy (0°/180° alternation) and no per-plant curvature or hinge tuning.}
  \label{fig:maize_baseline}
\end{figure}
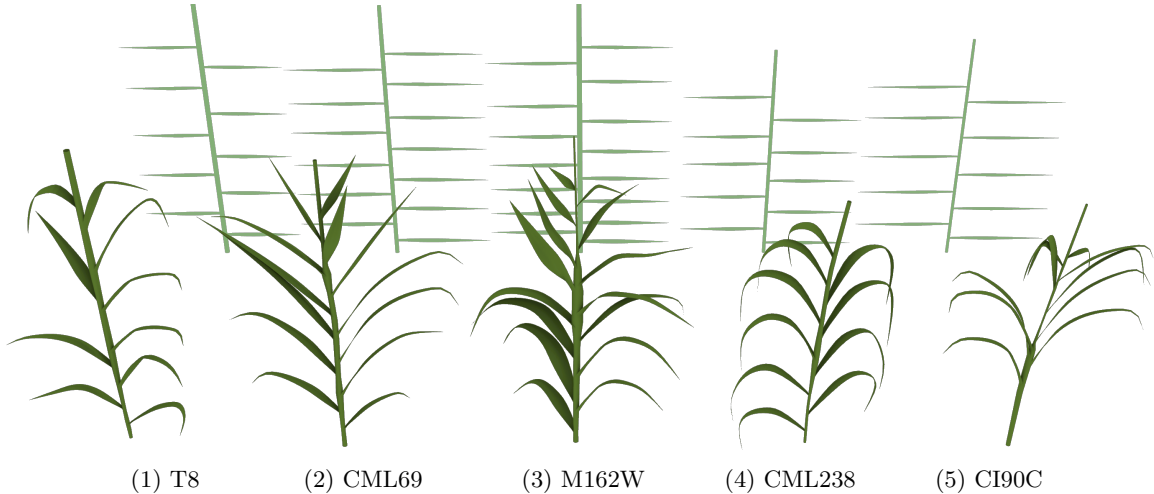

\begin{figure}[!t]
  \centering
  \includegraphics[width=0.8\linewidth]{Figures/cd_instance_vs_template.png}
  \caption{Symmetric CD for maize:
  LLM-Seeded Initial Template vs. Manual-Tuned 3D Model Instance. Manual tuning reduces the mean CD from
  0.102\,m to 0.021\,m (mean reduction $\approx$79\%).}
  \label{fig:cd_maize_template_instance}
\end{figure}

\subsubsection{Mung Bean Models}
\label{subsec:mungbean_models}

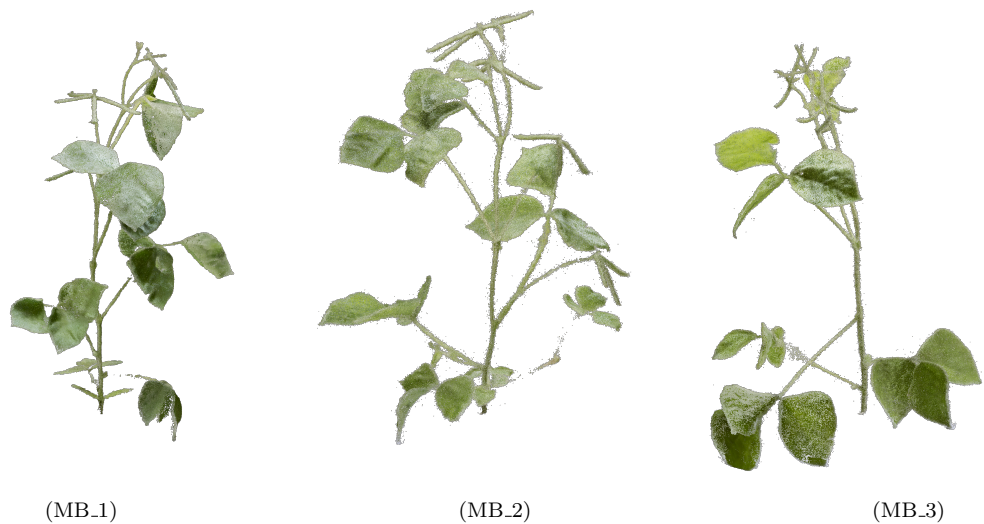
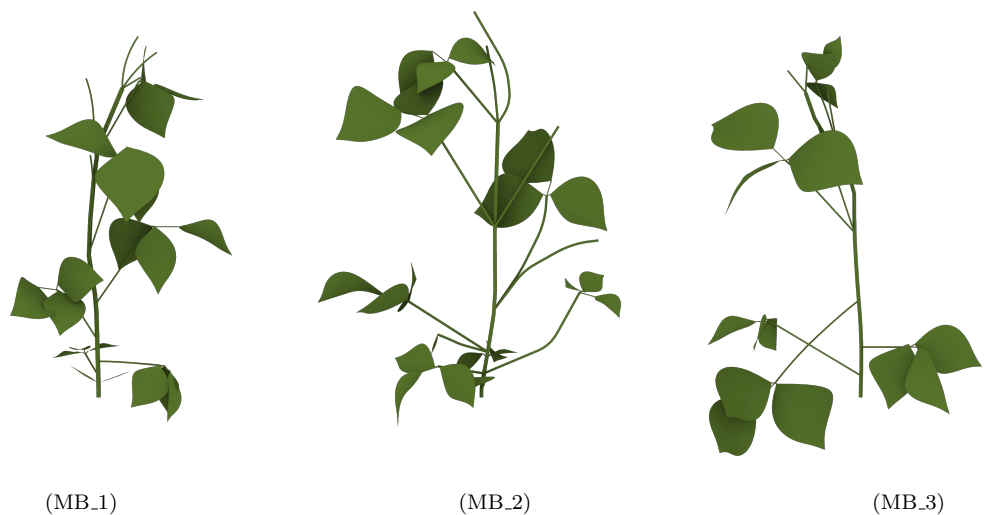
\begin{figure}[t!]
    \centering
    \begin{subfigure}{\textwidth}
        \centering
        \includegraphics[trim=0cm 13cm 0cm 13cm, clip, width=0.8\textwidth]{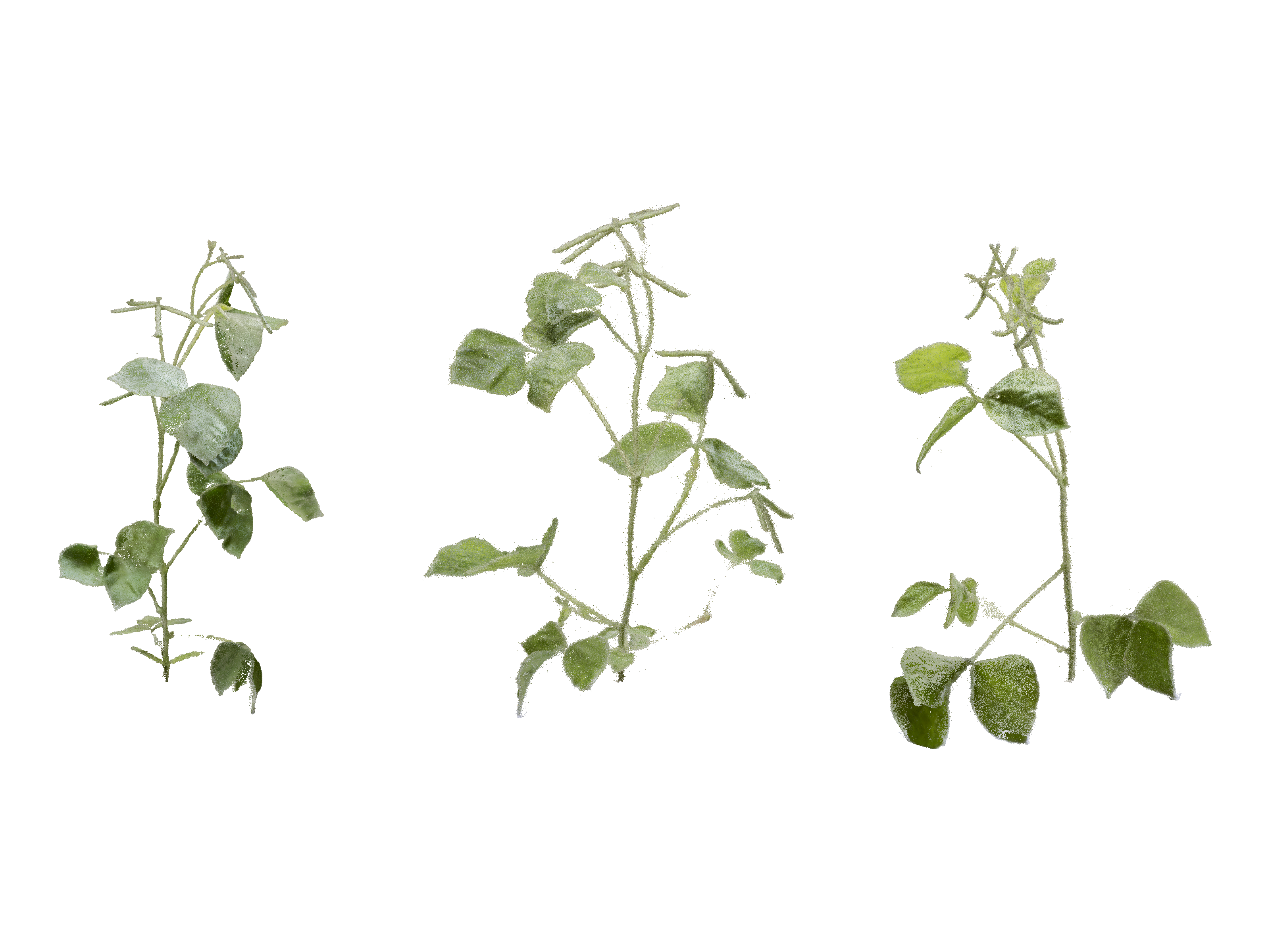}\\
        \setlength{\unitlength}{1cm}
        \begin{picture}(0,0)
            \put(-6.0,0.5){\scriptsize (MB\_1)}
            \put(-0.5,0.5){\scriptsize (MB\_2)}
            \put(5.0,0.5){\scriptsize (MB\_3)}
        \end{picture}
        \caption{Original mung bean point cloud data}
        \label{fig:mungbean_original}
    \end{subfigure}
    \begin{subfigure}{\textwidth}
        \centering
        \includegraphics[trim=0cm 13cm 0cm 13cm, clip, width=0.8\textwidth]{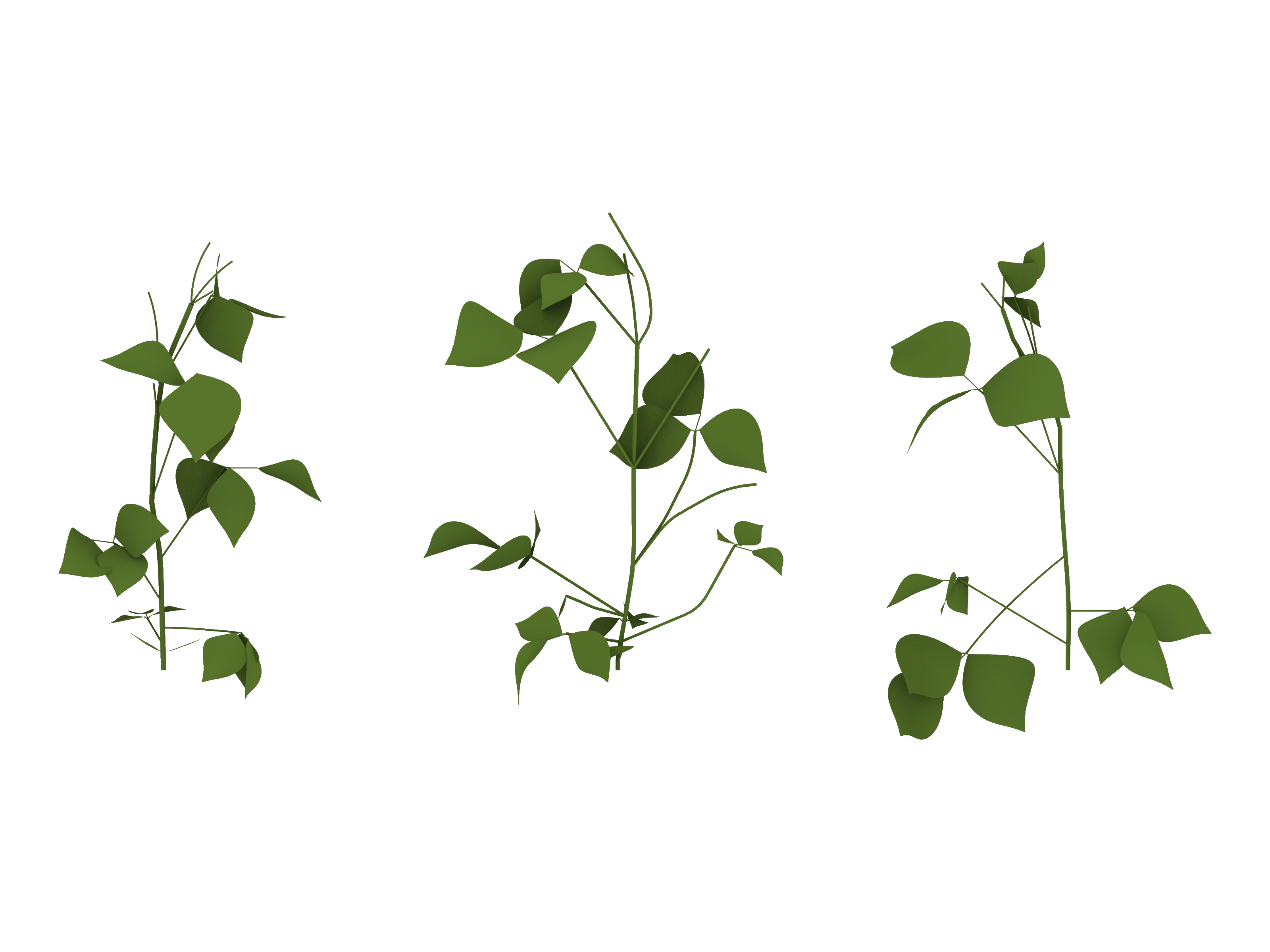}\\
        \setlength{\unitlength}{1cm}
        \begin{picture}(0,0)
            \put(-6.0,0.5){\scriptsize (MB\_1)}
            \put(-0.5,0.5){\scriptsize (MB\_2)}
            \put(5.0,0.5){\scriptsize (MB\_3)}
        \end{picture}
        \caption{Corresponding procedurally generated models}
        \label{fig:mungbean_procedural}
    \end{subfigure}
    \caption{Comparison of original point cloud data (a) and fitted procedural model instances (b) for three mung bean plants. The hierarchical compound leaf architecture (stalk→petiole→petiolule→leaflet) is accurately reproduced, demonstrating the framework's capability to represent complex multi-level branching structures characteristic of dicotyledonous species.}
    \label{fig:mungbean_fitted}
\end{figure}

Three mung bean plants were modeled to demonstrate the pipeline's capability to represent dicotyledonous compound leaf architectures. \figref{fig:mungbean_fitted} compares empirical point cloud data with fitted procedural models, illustrating the framework's capacity to capture hierarchical branching structures characteristic of trifoliate legumes.

\begin{figure}[t]
  \centering
  \includegraphics[width=0.8\linewidth]{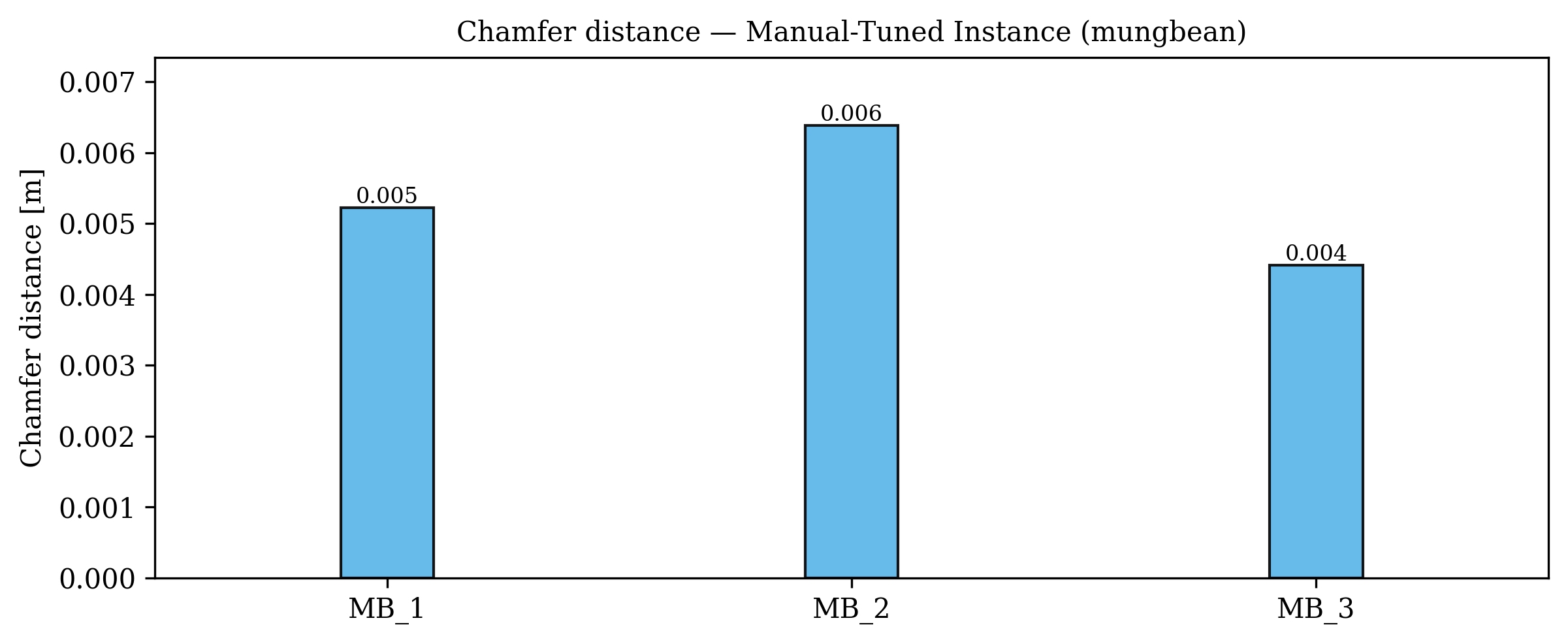}
  \caption{Symmetric CD between each fitted model instance and its point cloud for three mung bean plants (MB\_1–MB\_3).}
  \label{fig:mungbean_cd}
\end{figure}

We quantified the geometric fidelity for mung bean plants using the symmetric CD
between each fitted model instance and its point cloud (\figref{fig:mungbean_cd}).
CD values were \(0.005\), \(0.006\), and \(0.004\)~m for MB\_1–MB\_3, respectively
(mean \(\approx 0.005\)~m; SD \(\approx 0.001\)~m), indicating sub-centimeter agreement between the procedurally generated models and the reconstructions.

\clearpage

\subsection{Morphological Diversity}

Analysis across fitted models quantifies morphological diversity and validates biological realism. \tabref{tab:param_summary} summarizes parameter ranges. Plants demonstrate substantial diversity across architectural scales (6.4-fold range in first node height), geometric parameters (curvature spanning 10-10$^\mathrm{3}$cm bending radii), and morphological features (leaf lengths 4.5-102 cm, orientation covering full spherical range). All parameter ranges fall within biologically plausible values, validating morphological fidelity. Maize models extensively use hinge bending (10-35 hinge points per plant), while dicot models achieve complexity through hierarchical branching and leaflet orientation.

\begin{table}[h!]
\centering
\caption{Morphological parameter ranges across fitted plant models}
\label{tab:param_summary}
\small
\begin{tabular}{lcccccc}
\toprule
\textbf{Parameter} & \textbf{T8} & \textbf{CML69} & \textbf{M162W} & \textbf{CML238} & \textbf{CI90C} & \textbf{Unit} \\
\midrule
\multicolumn{7}{l}{\textit{Maize Plant Architecture}} \\
First node height & 10 & 12 & 7 & 8 & 45 & cm \\
Total height & 201 & 196 & 197 & 163 & 171 & cm \\
Number of leaves & 10 & 13 & 15 & 10 & 10 & -- \\
Max diameter & 42 & 52 & 70 & 40 & 40 & mm \\
Max curvature (\(\kappa\)) & 0.007 & 0.010 & 0.005 & 0.005 & 0.011 & rad/cm \\
Leaf length range & 60--80 & 50--102 & 20--85 & 50--78 & 25--102 & cm \\
Pitch angle range & 15--55 & 32--70 & 15--45 & 0--30 & 20--73 & deg \\
Hinges per leaf & 0--2 & 0--1 & 0--5 & 2--3 & 1--3 & -- \\
\midrule
\multicolumn{7}{l}{\textit{Mung Bean Architecture}} \\
& \textbf{MB\_1} & \textbf{MB\_2} & \textbf{MB\_3} & & & \\
Total leaflets & 21 & 20 & 15 & & & -- \\
Petiole length range & 3.0--16.0 & 11.0--20.5 & 4.5--19.8 & & & cm \\
Leaflet length range & 4.5--11.0 & 6.5--11.5 & 5.7--12.0 & & & cm \\
V-fold angle range & 5--20 & 10--20 & 5--20 & & & deg \\
\bottomrule
\end{tabular}
\end{table}

\section{Conclusions}
\label{sec:conclusion}
We presented FloraForge, an LLM-assisted procedural modeling framework that enables domain scientists to generate editable, analysis-ready 3D plant models without requiring expertise in geometric modeling. Users can progressively refine PPG that implement hierarchical B-spline surface representations with botanical constraints through iterative PR. The framework was validated across 8 plants spanning monocotyledonous (maize) and dicotyledonous (soybean, mung bean) architectures, demonstrating successful adaptation to fundamentally different morphological organizations through human-readable PDs. The continuous B-spline representation uniquely combines three essential properties: explicit parametric control for intuitive morphological editing, mathematical continuity ensuring smooth surfaces suitable for quantitative analysis, and direct compatibility with isogeometric analysis and computational simulation workflows. The dual-output format (STL meshes for visualization, SMESH parametric surfaces for analysis) addresses both graphics-oriented and computation-oriented applications.

Several future extensions would enhance the utility and automation of the framework. \textit{Species generalization} to additional crops (wheat, rice, sorghum) and growth forms (rosette plants, vining species, trees) would demonstrate the breadth of the LLM-assisted co-design approach. Each new species requires a single template creation phase via LLM PR, after which multiple instances are generated through PD parameter editing. \textit{Automated parameter extraction} from point clouds would eliminate the current manual fitting workflow (2-6 hours per plant), enabling high-throughput phenotyping. Potential approaches include geometric feature extraction with direct parameter mapping, gradient-based inverse procedural optimization, or learning-based methods where neural networks predict PD parameters from 3D scan input. \textit{Temporal modeling} via parameter interpolation across developmental stages would enable growth simulation and prediction, with time-series point cloud data constraining continuous growth trajectories. Finally, \textit{integration with simulation workflows}, including computational fluid dynamics for microclimate modeling, finite element analysis for lodging prediction, and global illumination for photosynthesis simulation, would demonstrate the analysis-ready nature of the B-spline representation for quantitative phenotyping and physics-based modeling applications.

By bridging the expertise gap between botanical domain knowledge and geometric modeling, FloraForge democratizes sophisticated 3D modeling capabilities for plant phenotyping, breeding, and precision agriculture research. The LLM-assisted co-design workflow transforms procedural modeling from a specialized skill into an accessible capability, enabling researchers to develop custom modeling tools tailored to specific scientific objectives. In the future, the efficiency and biological fidelity of the framework can be improved by integrating automated parameter optimization and temporal growth modeling. We hope that this work will be a starting point for advancing the role of human-AI collaboration in computational plant science.

\section*{Data Availability}
The code used to generate the procedural models in this work will be made publicly available upon acceptance.

\section*{Acknowledgements}
This work was supported by the AI Institute for Resilient Agriculture (USDA-NIFA 2021-67021-35329), NSF BTT-EAGER IOS-1842097, NSF 2412929/2412928 and Iowa State University's Plant Science Institute. It was also supported by NSF 2417510, 2412928, and 2309564 and USDA-NIFA 1032382 and 1032672 to Purdue.

\bibliographystyle{elsarticle-num-names} 
\bibliography{Refs}

\clearpage
\appendix

\renewcommand{\thefigure}{A.\arabic{figure}}
\setcounter{figure}{0}
\setcounter{table}{0}

\begin{figure}[t!]
    \centering
    \scriptsize  
    \begin{minipage}[t]{0.48\textwidth}
    \centering
    \textbf{Plant (4) CML238}
    \vspace{0.2cm}
    
    \begin{verbatim}
# Plant structure
outfile: out/maize_CML238.stl
seed: 2025

# Culm geometry (cm)
node-z: [8, 20, 34, 47, 60, 74,
         83, 96, 109, 127]

# Diameter profile (mm)
stem-diameter-mm: [15, 15, 15, 15,
                   24, 40, 33, 25,
                   25, 25]

# Culm curvature
stalk-kappa: [0.00, 0.00, 0.005,
              0.0, 0.00, 0.00,
              0.00, 0.00, 0.0,
              0.005]
stalk-bend-az-deg: [0, 0, 318, 0,
                    0, 0, 0, 0,
                    0, 318]

# Phyllotaxy (degrees)
leaf-az-deg: [116, 315, 115, 320,
              127, 320, 110, 330,
              115, 110]
leaf-pitch-deg: [0, 15, 15, 25, 30,
                 25, 30, 25, 30, 30]

# Leaf morphology defaults
leaves:
  globals:
    ctrl_u: 9
    ctrl_v: 5
    camber: 0.018
    camber_pow: 1.4
    width_pow: 1.20
    \end{verbatim}
    \end{minipage}%
    \hfill
    \begin{minipage}[t]{0.48\textwidth}
    \centering
    \textbf{Plant (5) CI90C}
    \vspace{0.2cm}
    
    \begin{verbatim}
# Plant structure
outfile: out/maize_CI90C.stl
seed: 2025

# Culm geometry (cm)
node-z: [45, 57, 65, 70, 75, 96,
         113, 119, 129, 135]

# Diameter profile (mm)
stem-diameter-mm: [25, 35, 38, 40,
                   38, 10, 8, 15,
                   9, 13]

# Culm curvature
stalk-kappa: [0.002, 0.001, 0.002,
              0.0, 0.011, 0.002,
              0.00, 0.00, 0.0,
              0.00]
stalk-bend-az-deg: [280, 280, 180,
                    0, 288, 200,
                    0, 0, 0, 0]

# Phyllotaxy (degrees)
leaf-az-deg: [136, 272, 118, 140,
              273, 274, 80, 272,
              98, 288]
leaf-pitch-deg: [35, 40, 50, 40, 48,
                 45, 73, 35, 40, 20]

# Leaf morphology defaults
leaves:
  globals:
    ctrl_u: 9
    ctrl_v: 5
    camber: 0.018
    camber_pow: 1.4
    width_pow: 1.20
    \end{verbatim}
    \end{minipage}
    \begin{minipage}{\textwidth}
    \centering
    \textbf{Per-leaf parameter overrides with localized bending}
    \vspace{0.2cm}
    
    \begin{minipage}[t]{0.48\textwidth}
    \begin{verbatim}
# CML238 - Leaf 3
nodes:
  3:
    length: 70
    width: 9.5
    bend: 0.35
    leaf-bend-hinges:
      - {u0: 0.1, angle-deg: 30,
         axis: B, smooth: 0.02}
      - {u0: 0.35, angle-deg: 20,
         axis: B, smooth: 0.02}
      - {u0: 0.85, angle-deg: 20,
         axis: B, smooth: 0.02}
    \end{verbatim}
    \end{minipage}%
    \hfill
    \begin{minipage}[t]{0.48\textwidth}
    \begin{verbatim}
# CI90C - Leaf 3
nodes:
  3:
    length: 91
    width: 8.5
    bend: 0.15
    leaf-bend-hinges:
      - {u0: 0.82, angle-deg: 90,
         axis: B, smooth: 0.03}
    \end{verbatim}
    \end{minipage}
    \end{minipage}
    \caption{Comparison of Plant Descriptors for two morphologically 
        distinct maize plants from \figref{fig:maize_fitted}. \textbf{Top:} 
        Plant-level structural parameters. CML238 exhibits gradual internode 
        spacing (8--127~cm) with minimal curvature, while CI90C shows elevated base height (first node at 45~cm) and pronounced culm bending ($\kappa=0.011$~rad/cm). \textbf{Bottom:} Per-leaf parameter specifications. CML238 leaf 3 contains three localized hinges for complex sigmoidal curvature, whereas CI90C leaf 3 uses a single dramatic 90° bend at $u=0.82$. The hierarchical structure 
        (global defaults + node-specific overrides) reduces parameter space while maintaining biological realism. All parameters are human-readable and directly editable without programming knowledge.}
    \label{fig:yaml_configs}
\end{figure}

\end{document}